\pdfoutput=1
\documentclass[11pt]{article}

\usepackage[final]{acl}

\usepackage{times}
\usepackage{latexsym}

\usepackage[T1]{fontenc}
\usepackage{algorithm}
\usepackage{algpseudocode}
\usepackage{amsmath}
\usepackage{amssymb}% http://ctan.org/pkg/amssymb
\usepackage{pifont}% http://ctan.org/pkg/pifont
\newcommand{\cmark}{\ding{51}}% Check mark
\newcommand{\xmark}{\ding{55}}% Cross mark

\usepackage[utf8]{inputenc}

\usepackage{microtype}

\usepackage{inconsolata}

\usepackage{graphicx}
\graphicspath{ {./images/} }

\title{Graph RAG-Tool Fusion}

\author{
 \textbf{Elias Lumer}, 
 \textbf{Pradeep Honaganahalli Basavaraju}, 
 \textbf{Myles Mason}, \\
 \textbf{James A. Burke and} 
 \textbf{Vamse Kumar Subbiah} \\
 \textit{PricewaterhouseCoopers, AI \& EmTech Teams} \\ 
 \small{\textbf{Correspondence:} \href{mailto:elias.lumer@pwc}{elias.lumer@pwc}} 
}

\begin{document}
\maketitle
\begin{abstract}
Recent developments in retrieval-augmented generation (RAG) for selecting relevant tools from a tool knowledge base enable LLM agents to scale their complex tool calling capabilities to hundreds or thousands of external tools, APIs, or agents-as-tools. However, traditional RAG-based tool retrieval fails to capture structured dependencies between tools, limiting the retrieval accuracy of a retrieved tool's dependencies. For example, among a vector database of tools, a "get stock price" API requires a "stock ticker" parameter from a "get stock ticker" API, and both depend on OS-level internet connectivity tools. In this paper, we address this limitation by introducing Graph RAG-Tool Fusion, a novel plug-and-play approach that combines the strengths of vector-based retrieval with efficient graph traversal to capture all relevant tools (nodes) along with any nested dependencies (edges) within the predefined tool knowledge graph. We also present ToolLinkOS, a new tool selection benchmark of 573 fictional tools, spanning over 15 industries, each with an average of 6.3 tool dependencies. We demonstrate that Graph RAG-Tool Fusion achieves absolute improvements of 71.7\% and 22.1\% over naïve RAG on ToolLinkOS and ToolSandbox benchmarks, respectively (mAP@10). 

\end{abstract}

\section{Introduction}

Recent advancements in Large Language Models (LLMs) enable agents to dynamically interact with external tools, APIs, or agents-as-tools, such as research gathering, software development lifecycle, or customer service. With breakthroughs in advanced retrieval-augmented generation (RAG) for tool selection, which store and retrieve relevant tools from a vector database and equip to an agent on inference-time, LLM agents can scale their tool calling capacity to hundreds of thousands of tools \cite{chen2024reinvoketoolinvocationrewriting, lumer2024toolshedscaletoolequippedagents}.

Despite traditional RAG approaches for tool selection that excel in retrieving tools based on unstructured, semantic relationships, they face limitations in representing structured dependencies between tools. For example, a “\textit{restaurant\_reservation}” tool may depend on “\textit{get\_current\_location}” and “\textit{get\_current\_datetime}” tools to provide function parameters such as location, date, or time. Moreover, existing tool retrieval benchmarks fail to address such dependencies, making it difficult to evaluate real-world scenarios involving interdependent tools.

\begin{figure*}[t]
  \centering
  \includegraphics[width=\textwidth]{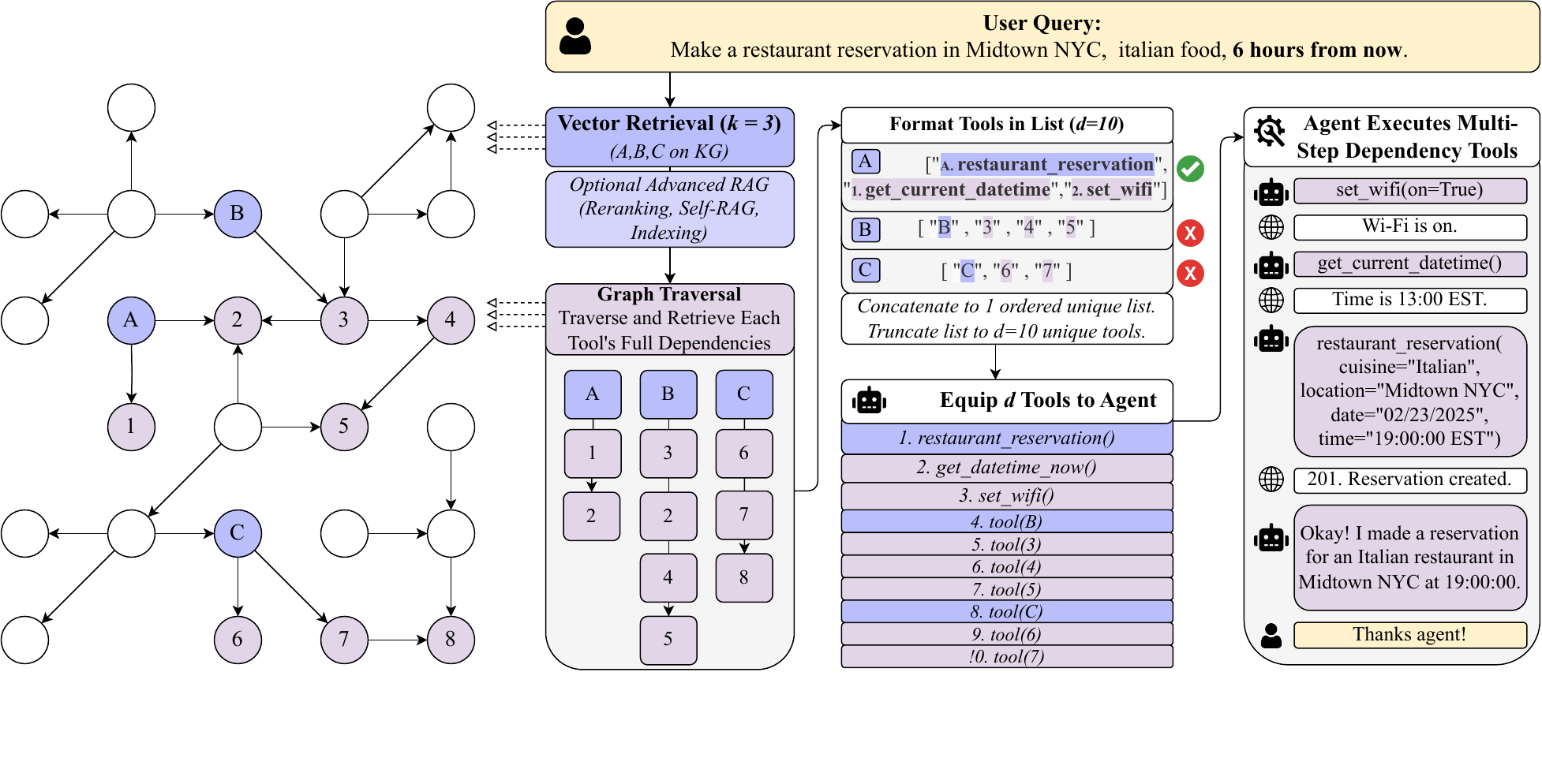} 
  \caption{The Graph RAG-Tool Fusion approach. After a user query, retrieve initial k tools through vector search (left KG, in letters) and traverse through each of its dependencies (left KG, in numbers) in the knowledge graph (pre-indexed). Once all tools are gathered, the tools are ordered in a unique list starting with the top vector tool followed by its dependencies, then the next vector tool and its dependencies, and so on, limited to final top-K tools. Finally, equip the top-K retrieved tools to the agent to solve the multi-step user query.}
  \label{fig:graph_rag_tool_fusion_workflow}
\end{figure*}

In this paper, we introduce Graph RAG-Tool Fusion, a novel plug-and-play approach that combines the strengths of vector-based retrieval with efficient graph traversal to capture all relevant tools (nodes) along with any nested tool dependencies (edges) within the predefined tool knowledge graph (Fig~\ref{fig:graph_rag_tool_fusion_workflow}). During inference, a first-pass vector search retrieves initial \textit{top-k} relevant tools, followed by graph traversal to retrieve each tool’s \textit{d} dependencies up to a final \textit{top-K} tools, and finally equipping the tools to an agent to solve the user question.

To address the lack of benchmarks focusing on inter-tool dependencies, we present ToolLinkOS, a dataset comprising 573 fictional tools spanning over 15 industries, each with an average of 6.3 total dependencies (Table~\ref{tab:tool_comparison}). We demonstrate that Graph RAG-Tool Fusion significantly outperforms naïve RAG, achieving absolute improvements of 71.7\% and 21.1\% on the ToolLinkOS and ToolSandbox benchmarks, respectively (mAP@10).

\section{Related Works}
\subsection{Advanced Retrieval-Augmented Generation}
Advanced retrieval-augmented generation (RAG) methods enhance document retrieval through pre-processing techniques (e.g., sliding window chunking, context-enriched chunking, small-to-big, parent-child chunking, and reverse HyDe chunking) \cite{anthropic_contextual_2024,setty_improving_2024,yang_advanced_2023}, intra-retrieval strategies (e.g., query rewriting, decomposition, and transformation) \cite{ma_query_2023,tang_multihop-rag_2024,trivedi_interleaving_2023,yao_react_2023,khattab_demonstrate-search-predict_2023,joshi_reaper_2024,xu_rewoo_2023,zheng_take_2024}, and post-retrieval approaches (e.g., reranking and corrective RAG) \cite{raudaschl_forget_2023,asai_self-rag_2023,yan_corrective_2024,sun_is_2023}. While Graph RAG is within the advanced RAG family, traditional RAG relies on vector databases, whereas Graph RAG uses knowledge graphs and vector retrieval \cite{gao2024retrievalaugmentedgenerationlargelanguage}. Retrieved documents from Graph RAG can also be concatenated with vector search results \cite{sarmah2024hybridragintegratingknowledgegraphs,raudaschl_forget_2023}. Two-pass document retrieval strategies like small-to-big and parent-child chunking \cite{langchain_parent_document_retriever_2023}, which map smaller chunks to larger chunks, require maintaining child-parent ID mappings and performing inefficient vector retrieval, whereas structured knowledge graphs naturally manage the indexing, mapping, and retrieval during inference-time graph traversal. Our Graph RAG-Tool Fusion is a novel plug-and-play approach extending Graph RAG from document retrieval to scalable tool selection for agents, leveraging advanced RAG strategies for the initial vector retrieval.

\subsection{Knowledge Graphs and LLMs} Integrating knowledge graphs (KGs) and LLMs enhances RAG by leveraging structured relationships between document chunks and entities to improve reasoning, retrieval, Q\&A, and summarization tasks \cite{peng2024graphretrievalaugmentedgenerationsurvey, neo4j_graphrag_2024}. Common approaches utilize an LLM or agent to decompose queries, iteratively explore a knowledge graph, extract relevant subgraphs, and solve multi-hop queries \cite{sun2024thinkongraphdeepresponsiblereasoning, li2024decodinggraphsfaithfulsound, jin2024graphchainofthoughtaugmentinglarge,hu2024graggraphretrievalaugmentedgeneration}. Combining various data structures such as graphs, chunks, tables, or retrievers can further tailor queries to correct document sources \cite{li2024structragboostingknowledgeintensive,sarmah2024hybridragintegratingknowledgegraphs}. Knowledge graphs improve LLM long-term memory \cite{gutiérrez2024hipporagneurobiologicallyinspiredlongterm}, text-to-SQL and text-to-Cypher translation tasks \cite{sequeda2023benchmarkunderstandroleknowledge,ozsoy2024text2cypherbridgingnaturallanguage}, and domain-specific approaches \cite{wu2024medicalgraphragsafe}. Graph learning approaches encode structural and relational information from graphs to support node classification, link prediction, and graph-level analysis \cite{he2024gretrieverretrievalaugmentedgenerationtextual,jiang2023diffkgknowledgegraphdiffusion,zaratiana2024grapherstructureawaretexttographmodel}.

Unlike these approaches, Graph RAG-Tool Fusion uses a knowledge graph for tool selection rather than document Q\&A or schema creation, avoiding reliance on graph learning, automatic schema generation, or text-to-Cypher prompting.

\subsection{Tool Selection or Retrieval}
Tool selection aims to retrieve a subset of relevant tools from a large corpus of tools. Baseline approaches utilize lexical term matching for storage and retrieval. Recently, state-of-the-art retriever-based and LLM-based approaches rely on traditional RAG and vector databases for tool selection. Retriever-based methods utilize neural networks to learn the semantic relationships between user queries and tools \cite{anantha_protip_2023,chen2024reinvoketoolinvocationrewriting,lumer2024toolshedscaletoolequippedagents,moon_efficient_2024}. LLM-based methods use an LLM or agent to plan, retrieve, and select relevant tools \cite{yuan_easytool_2024,li_api_bank_2023,du_anytool_2024}. Graph-based methods aim to plan out multi-hop queries with available APIs, but do not consider direct or indirect tool dependencies and benefits of vector retrieval \cite{liu2024toolnetconnectinglargelanguage,liu2023controlllmaugmentlanguagemodels,zhang2023graphtoolformerempowerllmsgraph}.

Our approach, Graph RAG-Tool Fusion, falls within retriever-based tool selection, but introduces a novel use of knowledge graphs to enhance retrieval. To our knowledge, it is the first method to combine knowledge graphs with vector retrieval to improve tool selection, enabling the efficient indexing, maintenance, and retrieval of structured direct or indirect dependencies between tools without relying on query planning strategies.

\section{Graph RAG-Tool Fusion}
Graph RAG-Tool Fusion is a novel plug-and-play approach for tool selection that combines the strengths of vector search with efficient knowledge graph traversal, requiring no model fine-tuning. Initially, Graph RAG-Tool Fusion retrieves relevant subgraphs of tools by performing a first-pass vector search to identify relevant tools, followed by graph traversal to retrieve all direct or indirect dependencies of those tools. Graph RAG-Tool Fusion captures both semantic relevance and relationships between tools and user queries, addressing the limitations of purely vector-based approaches. The graph index is built using a predefined schema to model tool nodes and edges.

\subsection{Graph Indexing of Tools}\label{subsec:graph_indexing}
From a large corpus of \textit{M} total tools, APIs, or agents-as-tools, transform each \textit{t} tool into a schema model, characterizing it as either a regular tool or a core tool (node), with an optional list of any direct or indirect dependencies (edge) to other tools (regular or core).

\subsubsection{Core Tool (Node Type)}\label{subsubsec:core_tool}
\begin{figure}[t]
  \centering
  \includegraphics[width=1\columnwidth]{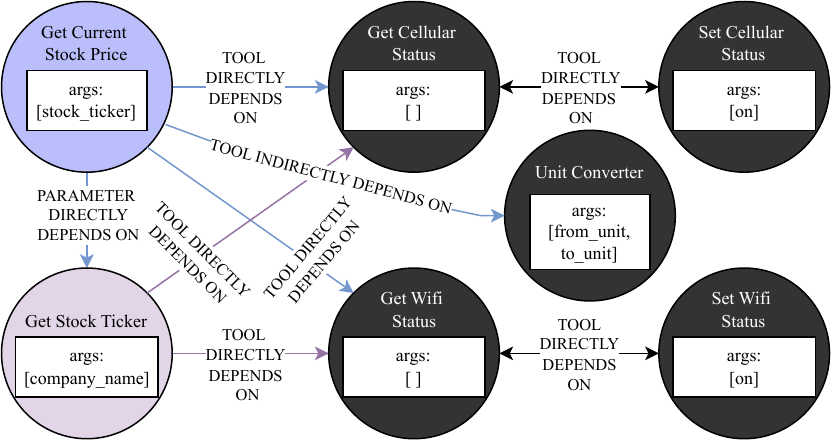} %trim=0cm 1.25cm 0cm 1.25cm, clip=true]
  \caption{Graph RAG-Tool Fusion's tool retrieval process detailing the predefined KG schema. The "Get Current Stock Price" tool is retrieved through vector search, then tool dependencies (e.g. "Get Stock Ticker") are retrieved through graph traversal, with schema relationships such as "parameter directly depends on."}
  \label{fig:graph_rag_tool_fusion_tool_selection_process}
\end{figure}

A core tool represents a reusable function that is a typical dependency of other functions, usually invoked by agents before using regular tools. For example, a function that returns the current date "\textit{get\_current\_date}" can be used by other functions that have dates in their parameters. Core tools can also have dependencies on core or regular tools, as regular tools do. More examples include OS-related tools such as "\textit{set\_wifi\_status}", "\textit{set\_cellular\_status}", or "\textit{unit\_converter}" (See Figure~\ref{fig:graph_rag_tool_fusion_tool_selection_process} and Appendix~\ref{sec:app_toolink_core}, Figure~\ref{fig:toollinkos_core}).

\subsubsection{Regular Tool (Node Type)}\label{subsubsec:regular_tool}
A regular tool represents a tool, API, or agent-as-tool that acts as a non-utility tool an agent can use. This tool often has dependencies on other tools (\textit{core} or \textit{regular}). For example, a "\textit{get stock price}" tool with a \textit{ticker} parameter relies on knowing the stock ticker, such as when a user asks for the current stock price of "Apple" (See Figure~\ref{fig:graph_rag_tool_fusion_tool_selection_process} and Appendix~\ref{sec:app_toolink_regular}, Figure~\ref{fig:toollinkos_regular}).

\subsubsection{Tool Relationships (Edges)}\label{subsubsec:tool_relationships}
Relationships between tool nodes represent four primary dependency types: \textit{tool directly depends on}, \textit{parameter directly depends on}, \textit{tool indirectly depends on}, and \textit{parameter indirectly depends on}. 
In the schema format, complementary labels of the relationship, \textit{reason} and \textit{parameter\_name}, explain why the dependency exists between the two tools and which parameter (if any) is dependent.

\paragraph{Tool directly depends on.} A tool requires another tool to operate. For example, tools such as \textit{"set\_wifi\_on"} or \textit{"set\_cellular\_on"} are essential for some tools to operate to receive a valid response.

\paragraph{Tool indirectly depends on.} A tool benefits from another tool but does not strictly require it. For instance, a \textit{"restaurant reservation"} tool might benefit from weather information (\textit{"get\_weather"}) but can function without it. 

\paragraph{Parameter directly depends on.} A parameter must be obtained from another tool before the primary tool can run. For example, a \textit{"product\_info"} tool requires a \textit{"product\_id"} parameter to proceed, which the \textit{"get\_product\_id"} tool can provide.

\paragraph{Parameter indirectly depends on.} A parameter may depend on additional context but only if explicitly required by the user input. For instance, the date \textit{"tomorrow"} requires the current date tool to compute the relevant date, but specifying \texttt{12/25/2025} does not. 

\subsection{Tool Retrieval}
As described in Algorithm~\ref{alg:graph_rag_tool_fusion}, Graph RAG-Tool Fusion begins by optionally transforming the input query using an advanced RAG transformation. It then retrieves an initial \textit{top-k} list of tools through vector search, optionally reranks the results, and performs a graph traversal using depth-first search to identify all dependencies (limit \textit{d-limit} per tool) for each tool in the retrieved list. Once all tools are gathered, the tools are ordered in a unique list starting with the top vector tool followed by its dependencies, then the next vector tool and its dependencies, and so on, limited to \textit{final top-K} tools.

\renewcommand{\algorithmicrequire}{\textbf{Input:}}
\renewcommand{\algorithmicensure}{\textbf{Output:}}

\begin{algorithm}[t]
\small
\caption{Graph RAG-Tool Fusion Retrieval}
\label{alg:graph_rag_tool_fusion}
\begin{algorithmic}[1]
\Require Vector similarity function $\tt{VectorSearch}$, \par Input query $q$, Vector DB $VDB$, \par
       Initial Vector top-k $top\_k$, \par
       Reranked Vector top-k $rerank\_top\_k$, \par
       Final top-K $final\_top\_K$, \par
       Graph index $KG$, \par
       Depth First Search function $\tt{DFS}$, \par
       Tool Dependency Limit $d\_limit$, \par
       Query transformation function $\tt{QueryTransform}$, \par
       Reranking function $\tt{Rerank}$

\State Initialize $S_{\text{graph\_list}} \gets []$

\If{$\tt{QueryTransform}$ is defined}
    \State $q \gets \tt{QueryTransform}(q)$
\EndIf

\State $S_{\text{initial\_vector\_tools}} \gets \tt{VectorSearch}(q, VDB, top\_k)$

\If{$\tt{Rerank}$ is defined}
    \State $S_{\text{reranked\_vector\_tools}} \gets \tt{Rerank}(S_{\text{initial\_vector\_tools}}, rerank\_top\_k)$
\Else
    \State $S_{\text{reranked\_vector\_tools}} \gets S_{\text{initial\_vector\_tools}}$
\EndIf

\For{each $t \in S_{\text{reranked\_vector\_tools}}$}
    \If{$t \notin S_{\text{graph\_list}}$}
        \State Append $t$ to $S_{\text{graph\_list}}$
    \EndIf
    \For{each tool $d \in \tt{DFS}(t, KG)$ \textbf{up to} $d\_limit$}
        \If{$d \notin S_{\text{graph\_list}}$}
            \State Append $d$ to $S_{\text{graph\_list}}$
        \EndIf
    \EndFor
\EndFor

\State $S_{\text{final\_graph\_list}} \gets \text{Limit } S_{\text{graph\_list}} \text{ to } final\_top\_K$

\State \Return $S_{\text{final\_graph\_list}}$

\end{algorithmic}
\end{algorithm}

\subsection{Graph RAG-Tool Fusion Retrieval Accuracy Equation}\label{subsec:retrieval_equation}
We model the expected retrieval accuracy of \emph{Graph~RAG-Tool~Fusion} as follows. 
Let $k$ denote the number of tools initially retrieved through vector search, 
$d$ the cut-off for each tool's total dependencies, 
$K$ the final top-$K$ cut-off, 
and $N$ the total number of tools discovered including all dependencies. 
As shown in Equation~\ref{eq:grtf_retrieval_accuracy}, 
the final retrieval accuracy is given by summing the baseline vector-retrieval accuracy plus any additional accuracy contributed by graph traversal of tool dependencies, 
scaled by the fraction of retrieved tools that fit into the final top-$K$ limit:
\begin{equation}\label{eq:grtf_retrieval_accuracy}
\begin{split}
\mathbb{E}[\mathit{GRTF\ Retrieval}(k,d,K)]
  = \\ \mathbb{E}\bigl[\mathrm{Retrieval}_{\mathrm{vector}}(k)\bigr] 
  \quad +\ \\\mathbb{E}\bigl[\mathrm{KG}_{\mathrm{dep}}(k,d)\bigr]
    \;\times\; \min\!\Bigl(1, \tfrac{K}{N}\Bigr),
\end{split}
\end{equation}

\section{Dataset Construction}\label{dataset_construction}

\begin{table*}[t]
  \scriptsize
  \centering
  \resizebox{\textwidth}{!}{%
  \begin{tabular}{p{3.5cm}cccccc}
    \hline
      & \textbf{ToolLinkOS} & \textbf{ToolSandbox} & \textbf{ToolBench} & \textbf{ToolE} & \textbf{Seal-Tools} & \textbf{ComplexFuncBench} \\
    \hline
    Number of Tools & 573 & 34 & 16,464 & 199 & 4076 & 48 \\
    Number of Instances & 1,569 & 1,032 & 126,486 & 21,127 & 14,076 & 1,000 \\
    Tool Dependencies & \cmark & \cmark & \xmark & \xmark & \xmark & \xmark \\
    KG Schema & \cmark & \xmark & \xmark & \xmark & \xmark & \xmark \\
    Avg. Dependencies per Tool & 6.3 & 1.6 & N/A & N/A & N/A & N/A \\
    Real API Response & \xmark & \cmark & \cmark & \xmark & \xmark & \cmark \\
    \hline
  \end{tabular}
  }
  \caption{Comparison of ToolLinkOS with other tool selection or tool calling benchmarks  \cite{lu2024toolsandboxstatefulconversationalinteractive,qin2023toolllmfacilitatinglargelanguage,huang2024metatoolbenchmarklargelanguage,wu2024sealtoolsselfinstructtoollearning,zhong2025complexfuncbenchexploringmultistepconstrained}.}
  \label{tab:tool_comparison}
\end{table*}

\subsection{ToolLinkOS Design}
The goal of ToolLinkOS is to create a large number of tools that depend on a subset of other tools. ToolLinkOS serves as the leading benchmark for complex, multi-hop, dependency-rich tool selection. While the tools are not functional, to our knowledge, it is the first tool retrieval dataset with pre-defined dependencies within a knowledge graph schema. All references to company names or services in this dataset are entirely fictional and used solely for illustrative purposes. No endorsement or affiliation is implied.

\subsubsection{ToolLinkOS Tools}\label{subsubsec:toollinkos_tools}
Tools in ToolLinkOS are categorized into core tools and regular tools (See Sections~\ref{subsubsec:core_tool} and~\ref{subsubsec:regular_tool}). Core tools serve as reusable, utility-like APIs that other tools depend on to define their parameters or enable their functionality. ToolLinkOS tools span over 15 industries, encompassing 523 regular tools and 50 core tools. Furthermore, each tool has an average of 6.3 dependencies, ranging from direct or indirect relationships.

\subsection{ToolSandbox KG Conversion}\label{subsubsec:toolsandbox_kg_conversion}
ToolSandbox \cite{lu2024toolsandboxstatefulconversationalinteractive} is an open-source tool dataset focused on functional APIs with dependencies to other functions, designed for testing in an environment to evaluate LLMs' ability to select the right tools, configure parameters, and understand dependencies. It contains 34 Python-native functions and selected RapidAPI tools that manipulate the world state, monitored through milestones and environments. We indexed their tools into our Graph RAG-Tool Fusion schema, consisting of tool nodes and four relationship edges (see Section~\ref{subsec:graph_indexing} for details). The resulting ToolSandbox knowledge graph is fictional and not runnable (mirroring ToolLinkOS), supporting tool retrieval tasks, with potential for future enhancements.

\subsection{Dataset Comparisons}
Table~\ref{tab:tool_comparison} compares ToolLinkOS with other tool selection or tool calling benchmarks--ToolSandbox \cite{lu2024toolsandboxstatefulconversationalinteractive}, ToolBench \cite{qin2023toolllmfacilitatinglargelanguage}, ToolE \cite{huang2024metatoolbenchmarklargelanguage}, Seal-Tools \cite{wu2024sealtoolsselfinstructtoollearning}, and ComplexFuncBench \cite{zhong2025complexfuncbenchexploringmultistepconstrained}--based on total tool count, total instances, average tool dependencies, whether it contains a knowledge graph schema, and whether the functions contain real API responses. Unlike existing datasets, ToolLinkOS contains a predefined knowledge graph schema with a large quantity of tools and dependencies, to assess complex dependency-rich tool selection. ToolSandbox, while offering real API responses, includes fewer tool dependencies and lacks a ready-to-use knowledge graph. ToolBench, ToolE, and Seal-Tools, while offering large tool collections, do not include clear tool dependencies or knowledge graph schemas.

\section{Experimental Settings}
\subsection{Embedding and LLM Models}
For all experiments, we utilize Azure OpenAI's {\tt text-embedding-ada-002} (embedding) and Azure OpenAI's {\tt gpt-4o-2024-08-06} (LLM).
\subsection{Benchmark Datasets}
\subsubsection{Datasets}
As stated in Section~\ref{dataset_construction}, the datasets we use in the subsequent experiments are ToolLinkOS (our contribution tool selection dataset) and ToolSandbox (converted to our KG schema). ToolLinkOS has 573 tools and ToolSandbox has 33 tools (1 tool is a human response tool which we omit).
\subsubsection{Metrics}
The primary metric used to evaluate retrieval performance, on 1,569 (ToolLinkOS) and 1,032 (ToolSandbox) instances, is mean absolute precision (mAP) at 10, 20, and 30. Additionally, we report full retrieval metrics for both benchmarks, including normalized discounted cumulative gain (nDCG) and recall at 10, 20, and 30, in Appendix \ref{sec:app_retrieval_performance_eval}, \ref{tab:full_retriever_comparison}.

\begin{table*}
  \scriptsize
  \centering
  \resizebox{\textwidth}{!}{%
  \begin{tabular}{p{2cm}lccc}
    \hline
    \textbf{Dataset} & \textbf{Retriever} & \textbf{mAP@10} & \textbf{mAP@20} & \textbf{mAP@30} \\
    \hline
                     & Lexical Search
                     & 0.215 & 0.217 & 0.217 \\
                     & Naïve RAG 
                     & 0.440 & 0.488 & 0.505 \\
    \textbf{ToolSandbox} 
                     & Hybrid RAG ($\alpha=0.8$) 
                     & 0.431 & 0.478 & 0.497 \\ (33 tools)
                     & Graph RAG-Tool Fusion ($k=3$), no RR
                     & \underline{0.521} & \underline{0.571} & \underline{0.571} \\
                     & \textbf{Graph RAG-Tool Fusion ($k=3$), w/ RR}
                     & \textbf{0.661} & \textbf{0.697} & \textbf{0.697} \\
\hline

                     & Lexical Search 
                     & 0.185 & 0.191 & 0.194 \\
                     & Naïve RAG 
                     & 0.210 & 0.216 & 0.217 \\
    \textbf{ToolLinkOS} 
                     & Hybrid RAG ($\alpha=0.8$) 
                     & 0.202 & 0.208 & 0.209 \\ (573 tools)
                     & Graph RAG-Tool Fusion ($k=3$), no RR
                     & \underline{0.856} & \underline{0.873} & \underline{0.873} \\ 
                     & \textbf{Graph RAG-Tool Fusion ($k=3$), w/ RR}
                     & \textbf{0.927} & \textbf{0.938} & \textbf{0.938} \\
    \hline
  \end{tabular}
  }
  \caption{Performance comparison on ToolSandbox \cite{lu2024toolsandboxstatefulconversationalinteractive} and our ToolLinkOS for different retrievers (BM25 \cite{robertson_probabilistic_2009}, Naïve RAG, Hybrid RAG (\(\alpha=0.8\)) \cite{gao2024retrievalaugmentedgenerationlargelanguage}, Graph RAG-Tool Fusion (\textit{k =} 3) without reranking ("no RR") and Graph RAG-Tool Fusion (\textit{k =} 3) with reranking initial vector retrieval tools ("w/ RR"). The table reports only mAP@10, mAP@20, and mAP@30; full metrics are available in Appendix~\ref{sec:app_retrieval_performance_eval}, Table~\ref{tab:full_retriever_comparison}.}
  \label{tab:condensed_retriever_comparison}
\end{table*}

\subsection{Tool Retrieval Baselines}

\paragraph{Lexical Search.}\label{subsec:lexical_rag}
As a baseline, lexical search (or text search) is evaluated to determine how much potential improvement arises from the semantic understanding of tools and user queries. Microsoft Azure's AI Search is employed for text search, leveraging Apache Lucene for indexing and querying, and the BM25 ranking algorithm for scoring results \cite{microsoft_azure_search_2023}.

\paragraph{Naïve RAG.}\label{subsec:naive_rag}
Naïve vector RAG search is examined as the primary comparison to Graph RAG-Tool Fusion. Semantic search is implemented using Microsoft Azure's AI Search \cite{microsoft_azure_search_2023}, which utilizes the HNSW algorithm. Parameter details are provided in Appendix~\ref{sec:app_ai_search_alg_params}, Figure \ref{fig:azure_ai_search_alg}.

\paragraph{Hybrid RAG.} 
Hybrid RAG, a weighted combination of lexical search and vector search, is also assessed \cite{kuzi2020leveragingsemanticlexicalmatching}. For this evaluation, $\alpha=0.8$ is used as the weight favoring vector search. Microsoft Azure's AI Search \cite{microsoft_azure_search_2023} is employed to implement hybrid search, maintaining consistency with the semantic and lexical configurations outlined in Sections~\ref{subsec:naive_rag} and Appendix~\ref{sec:app_ai_search_alg_params}, Figure~\ref{fig:azure_ai_search_alg}.

\subsection{Graph RAG-Tool Fusion Tool Retrieval}
\paragraph{Default Initial Retrieval.}
Graph RAG-Tool Fusion is evaluated at \(k=3\), initially retrieving the top-\(3\) tools based on hybrid search. The knowledge graph is implemented in Neo4j \cite{neo4j}.

\paragraph{Reranking Initial Retrieval.} 
Graph RAG-Tool Fusion is further assessed with advanced RAG methods applied to the initial vector retrieval. To enhance the relevance of retrieval, the top-\(k=3\) tools are reranked to prioritize the most relevant tool at the top during dependency traversal. 

\section{Experimental Results}
\subsection{Baseline Retrieval Performance}
As seen in Table \ref{tab:condensed_retriever_comparison}, BM25 has the lowest retrieval across baseline retrievers (mAP). Naïve RAG performs slightly better than Hybrid RAG in accuracy, both above lexical search, demonstrating the semantic relationships between queries and tools and indicating less of importance on lexical relationships.  These results hold true for both datasets, however, the difference between naïve RAG and Graph RAG-Tool Fusion is much larger in the ToolLinkOS benchmark. 

\subsection{Graph RAG-Tool Fusion Retrieval Performance}
Graph RAG-Tool Fusion performs significantly better than the highest baseline retriever (85.6\% to 21.0\% on ToolLinkOS and 52.1\% to 44.0\% on ToolSandbox, mAP@10). Additionally, Graph RAG-Tool Fusion with reranking outperforms standard Graph RAG-Tool Fusion by an absolute improvement of 7\% and 14\% on ToolLinkOS and ToolSandbox (mAP@10). 

\section{Discussions}
\subsection{Graph RAG-Tool Fusion}\label{subsec:grtf_discussion}
We evaluate the retrieval performance of Graph RAG-Tool Fusion compared to baselines including lexical search, naïve RAG, and Hybrid RAG, on the ToolLinkOS and ToolSandbox \cite{lu2024toolsandboxstatefulconversationalinteractive} benchmarks. Our results in Table~\ref{tab:condensed_retriever_comparison} provide evidence that the combination of graph and vector RAG (Graph RAG-Tool Fusion) significantly outperforms naïve RAG for tool retrieval with dependencies. ToolSandbox, with 33 tools and an average of 1.6 dependencies per tool, highlights the retrieval gap between naïve RAG and Graph RAG-Tool Fusion when contrasted with ToolLinkOS’s 573 tools and 6.3 average dependencies. Since tool dependencies are often semantically unrelated to the main tool, naïve RAG struggles to retrieve all relevant dependencies. In contrast, Graph RAG-Tool Fusion  leverages vector search for the main tool and traverses its structured dependencies during retrieval, outperforming naïve RAG.
\paragraph{Reranking.} We also provide evidence that reranking the initial \textit{k} tools retrieved by vector search increases retrieval accuracy by an absolute improvement of 7\% and 14\% on ToolLinkOS and ToolSandbox (mAP@10). Other advanced RAG strategies can be applied to the initial vector retrieval beyond reranking, including pre-retrieval, intra-retrieval, and post-retrieval techniques. For example, the user query can be rewritten, decomposed, or transformed before vectorization. 

\begin{figure}[t]
  \centering
  \includegraphics[width=1\columnwidth]{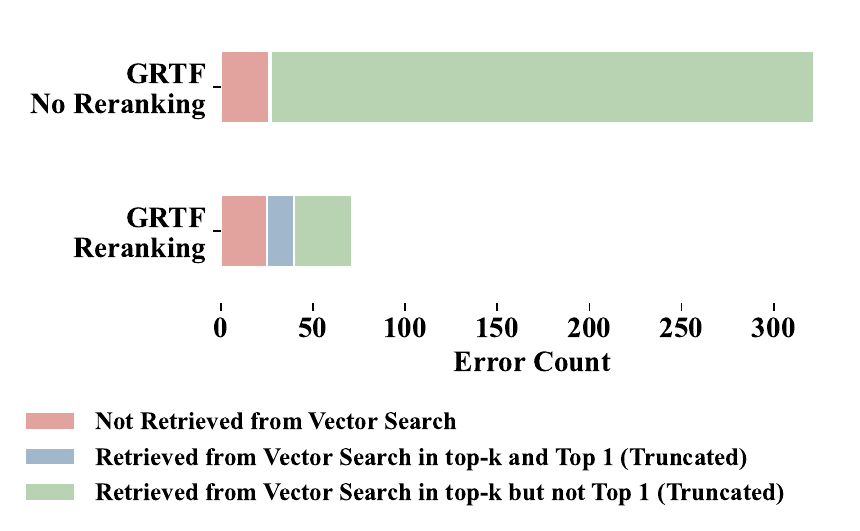} %trim=0cm 1.25cm 0cm 1.25cm, clip=true]
  \caption{Retrieval error rate of Graph RAG-Tool Fusion (ToolLinkOS), comparing the default GRTF to reranking with \textit{top-k} = 3. Red (top) indicates "not retrieved from vector search," blue (middle) represents "retrieved in \textit{top-k} and \textit{top-1} (truncated)," and green (bottom) is "retrieved in \textit{top-k} but not \textit{top-1} (truncated)."}
  \label{fig:retrieval_error_rate}
\end{figure}

\subsection{Retrieval Error Analysis}
In Figure \ref{fig:retrieval_error_rate}, we analyze retrieval errors to identify areas for improvement. The first error type (8.1\%) occurred when the correct primary tool was missing from the initial \textit{top-k} results (\textit{k=3}), affecting both reranked and non-reranked approaches equally.

The second and third errors stemmed from truncation of the final \textit{top-K} tools (see Section \ref{subsec:retrieval_equation}). The second error occurred when the correct tool was retrieved within \textit{top-k} but not as \textit{top-1}, affecting 91.6\% of GRTF (no RR) cases and 43.6\% of GRTF (RR). The third error, where the correct tool was ranked \textit{top-1} but excluded due to dependency truncation, accounted for <0.01\% of errors in GRTF (no RR) but 30\% in GRTF (RR). 

Improving the retrieval accuracy of Graph RAG-Tool Fusion involves primarily reducing errors within vector search by using more advanced RAG strategies. Notably, reranking reduced truncation errors by 52\% and improved accuracy by 7\% (ToolLinkOS, mAP@10).

\section{Conclusion}
As multi-agent systems become more complex with large-scale tools, APIs, or agents-as-tools, tools will often have dependencies on other tools, whether its utility functions, necessary tools to fill out parameters, or operating system-style tools. In this paper, we present Graph RAG-Tool Fusion, a novel plug-and-play approach that combines the strengths of vector-based retrieval with efficient graph traversal to capture all relevant tools (nodes) along with any nested dependencies (edges) within the predefined tool knowledge graph. We also present ToolLinkOS, a new tool selection benchmark of 573 tools, spanning over 15 industries, each with an average of 6 total tool dependencies. Furthermore, we demonstrate that Graph RAG-Tool Fusion significantly outperforms naïve RAG, achieving absolute improvements of 71.7\% and 22.1\% on the ToolLinkOS and ToolSandbox benchmarks, respectively (mAP@10).

\section{Limitations}
While Graph RAG-Tool Fusion is capable of significantly improving upon the tool retrieval accuracy of naïve RAG, pushing the tool selection and scalable agent community forward, it still has some limitations. Firstly, Graph RAG-Tool Fusion integrates vector search for first-pass retrieval, so the performance of the vector search retriever greatly affects the overall performance of Graph RAG-Tool Fusion. Secondly, Graph RAG-Tool Fusion requires a tool knowledge graph (from a schema) to be built for a set of 
\textit{M} tools. This work includes a manual process of creating the tool knowledge graph, which aimed to demonstrate its superior performance compared to naïve RAG (see Appendix \ref{sec:app_tool_generation_process}, Figure \ref{fig:tool_generation_process}). However, future work can build on this for automatic LLM-created tool knowledge graphs \cite{edge2024localglobalgraphrag, feng2024ontologygroundedautomaticknowledgegraph}. Lastly, while the tool relationship schema is broken down into 4 types (Section~\ref{subsubsec:tool_relationships}), Graph RAG-Tool Fusion's retrieval system does not prioritize certain relationships. Future work can prioritize relationships such as \textit{direct} ones over \textit{indirect} ones, if the dependency count and nested sub-graphs grow very large. 

\section{Ethical Considerations}
This paper was conducted in accordance with the ACM Code of Ethics. All results can be reproduced with the two publicly available datasets. Our data set includes disclaimers about referencing other companies in an illustrative, fictional manner. Furthermore, we did not hire additional researchers or workers for dataset creation and testing. Risks of our work include potential security vulnerabilities or system issues if OS-level tools are developed. Additionally, financial, health, and environmental tools need expert vetting despite our being created to the best of our knowledge of the domain.

\bibliography{references}

\appendix

\section{Tool Knowledge Graph Visualizations }\label{sec:app_graph_visualization}

\subsection{ToolSandbox Knowledge Graph Visualization}
\begin{figure*}[!h]
  \centering
  \includegraphics[width=\columnwidth]{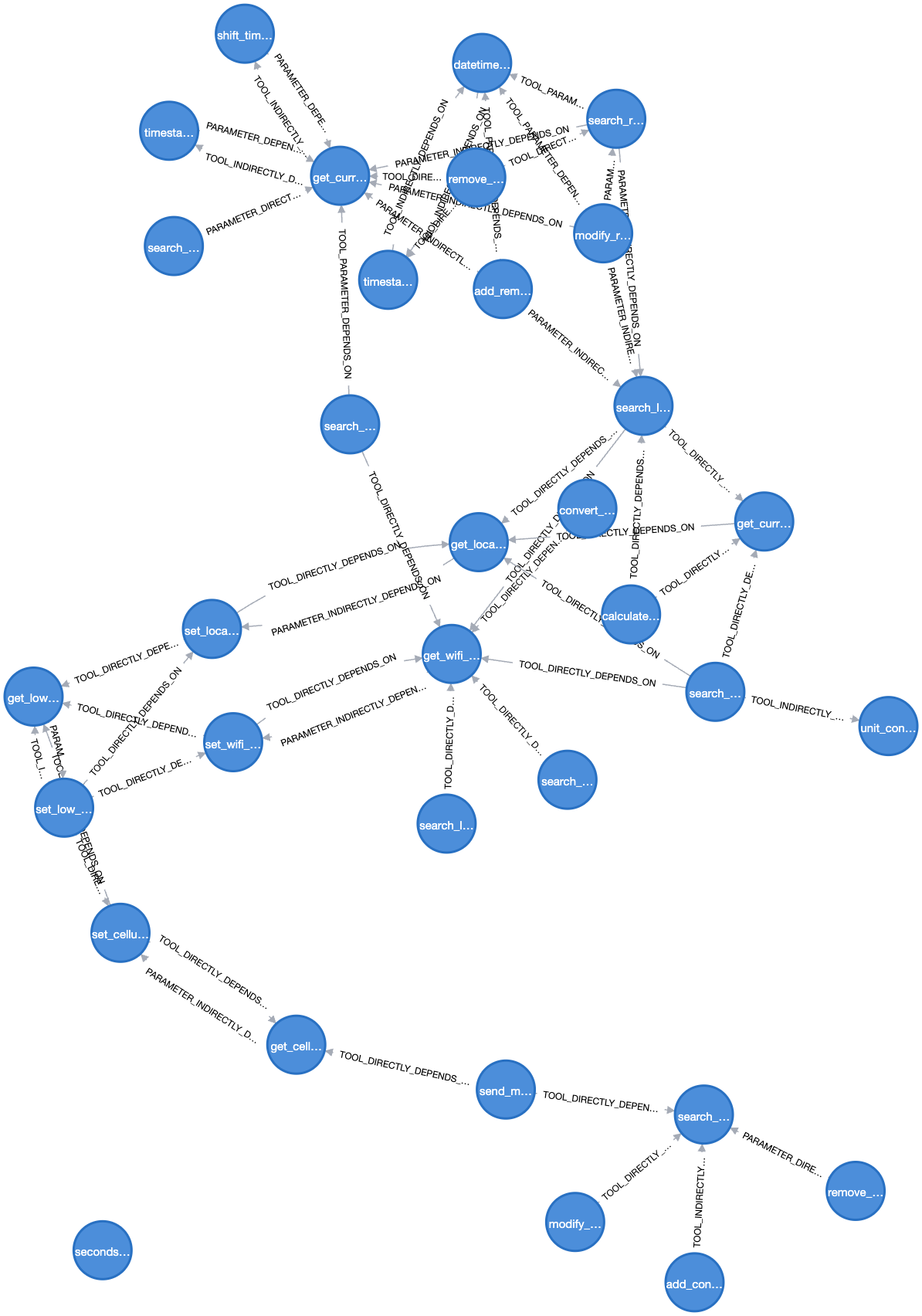} %trim=0cm 1.25cm 0cm 1.25cm, clip=true]
  \caption{Knowledge Graph of ToolSandbox, visualized in Neo4j, consisting of 33 nodes.}
  \label{fig:toolsandbox_visual}
\end{figure*}

In Figure~\ref{fig:toolsandbox_visual}, the full knowledge graph visualization of ToolSandbox is displayed. As explained in Section~\ref{subsubsec:toolsandbox_kg_conversion}, ToolSandbox \cite{lu2024toolsandboxstatefulconversationalinteractive} is converted into our Graph RAG-Tool Fusion KG Schema (See Section~\ref{subsec:graph_indexing} for schema details). 

\subsection{ToolLinkOS Knowledge Graph Visualization}
\begin{figure*}[!h]
  \centering
  \includegraphics[width=14cm]{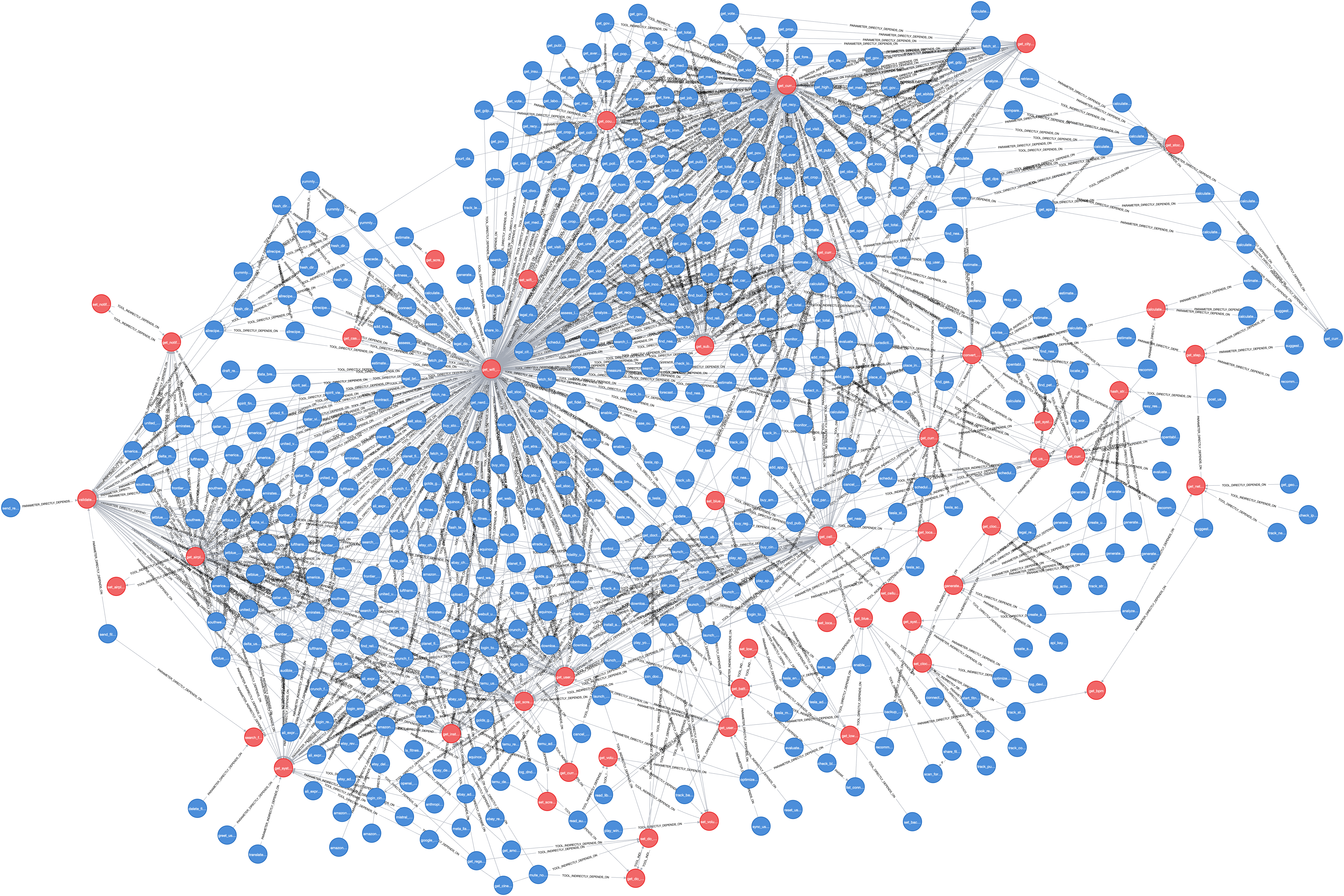} %trim=0cm 1.25cm 0cm 1.25cm, clip=true]
  \caption{Knowledge Graph of ToolLinkOS, visualized in Neo4j, consisting of 573 nodes.}
  \label{fig:toollinkos_visual}
\end{figure*}

In Figure~\ref{fig:toollinkos_visual}, the full knowledge graph visualization of our ToolLinkOS dataset is displayed. As explained in Section~\ref{subsubsec:toollinkos_tools},  the 573 tools are represented in a knowledge graph schema (See Section~\ref{subsec:graph_indexing} for schema details). Each tool has on average 6.3 dependencies (Table~\ref{tab:tool_comparison}). Useful core tools such as "\textit{get\_wifi\_status}", "\textit{get\_current\_time}", or "\textit{get\_current\_location}" have many dependencies, which is why the graph is partially centered around these set of popular core tools. 

\section{Full Error Analysis for Retrieval}
\label{sec:app_retrieval_performance_eval}
\begin{figure*}[!h]
  \centering
  \includegraphics[width=10cm]{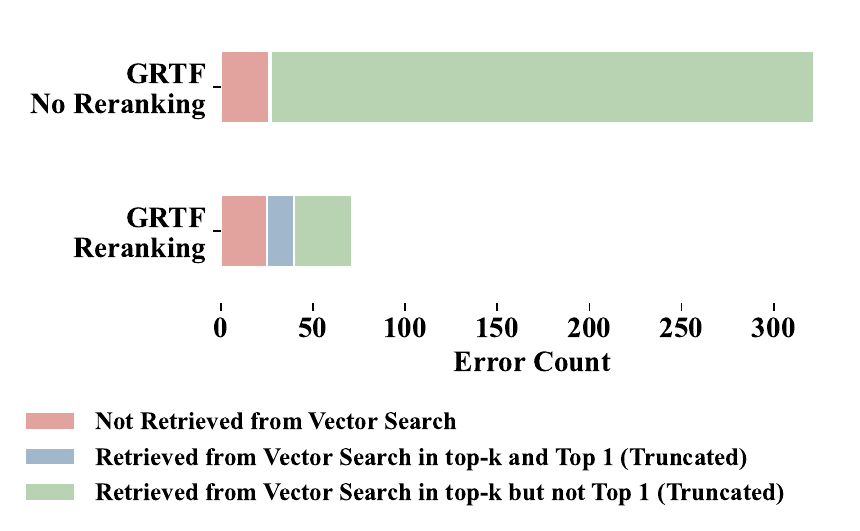} %trim=0cm 1.25cm 0cm 1.25cm, clip=true]
  \caption{Retrieval error rate of Graph RAG-Tool Fusion (ToolLinkOS), comparing a default implementation to reranking the initial vector retrieval \textit{top-k} = \textit{3}. Red (top) details the error "not retrieved from vector search," blue (middle) is "retrieved from vector search in top-k and top 1 (truncated)," and green (bottom) is "retrieved from vector search in top-k but not top 1 (truncated)."}
  \label{fig:retrieval_error_rate_ts}
\end{figure*}

\begin{figure*}[!h]
  \centering
  \includegraphics[width=10cm]{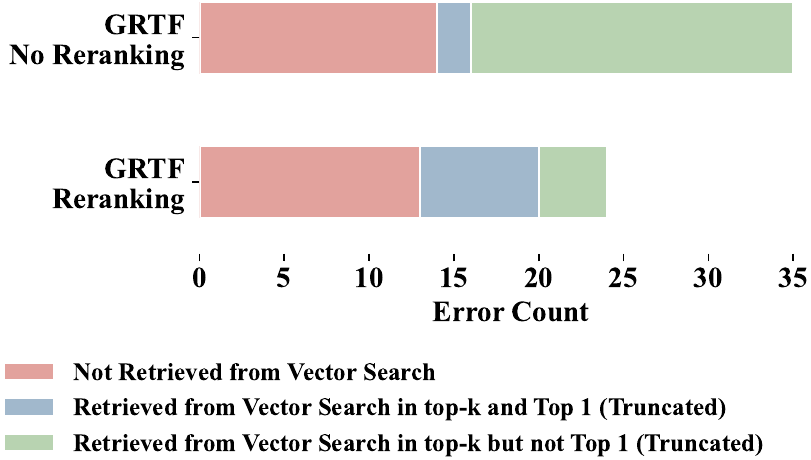} %trim=0cm 1.25cm 0cm 1.25cm, clip=true]
  \caption{Retrieval error rate of Graph RAG-Tool Fusion (ToolSandbox), comparing a default implementation to reranking the initial vector retrieval \textit{top-k} = \textit{3}. Red (top) details the error "not retrieved from vector search," blue (middle) is "retrieved from vector search in top-k and top 1 (truncated)," and green (bottom) is "retrieved from vector search in top-k but not top 1 (truncated)."}
  \label{fig:retrieval_error_rate_as}
\end{figure*}

Figures \ref{fig:retrieval_error_rate_ts} and \ref{fig:retrieval_error_rate_as} present a comprehensive retrieval error analysis for ToolLinkOS and ToolSandbox, respectively. 

\section{Full Retrieval Performance Evaluation}
\label{sec:app_retrieval_performance_eval}
The complete retrieval metrics are displayed in Table \ref{tab:full_retriever_comparison} including nDCG, mAP, and recall at 10,20,30 for benchmarks ToolSandbox and ToolLinkOS.

\begin{table*}[!h]
  \small
  \centering
  \resizebox{\textwidth}{!}{%
  \begin{tabular}{p{2cm}lccc|ccc|ccc}
    \hline
    \textbf{Dataset} & \textbf{Retriever} & \textbf{mAP@10} & \textbf{mAP@20} & \textbf{mAP@30} & \textbf{Recall@10} & \textbf{Recall@20} & \textbf{Recall@30} & \textbf{nDCG@10} & \textbf{nDCG@20} & \textbf{nDCG@30} \\
    \hline
                     & Lexical Search
                     & 0.215 & 0.217 & 0.217 & 0.301 & 0.307 & 0.307 & 0.303 & 0.306 & 0.306 \\
                     & Naïve RAG 
                     & 0.440 & 0.488 & 0.505 & 0.663 & \underline{0.850} & \underline{0.962} & \underline{0.606} & 0.682 & \underline{0.720} \\
    \textbf{ToolSandbox} 
                     & Hybrid RAG ($\alpha=0.8$) 
                     & 0.431 & 0.478 & 0.497 & \underline{0.665} & \textbf{0.854} & \textbf{0.968} & 0.592 & 0.669 & 0.708 \\ (33 tools)
                     % "no RR" is ~ 10% less
                     & Graph RAG-Tool Fusion ($k=3$), no RR
                     & \underline{0.521} & \underline{0.571} & \underline{0.571} & 0.598 & 0.833 & 0.833 & 0.601 & \underline{0.692} & 0.692 \\
                     % With RR = actual graph_rag_search_rerank numbers
                     & \textbf{Graph RAG-Tool Fusion ($k=3$), w/ RR}
                     & \textbf{0.661} & \textbf{0.697} & \textbf{0.697} & \textbf{0.704} & 0.833 & 0.833 & \textbf{0.740} & \textbf{0.795} & \textbf{0.795} \\
\hline

                     & Lexical Search 
                     & 0.185 & 0.191 & 0.194 & 0.253 & 0.293 & \underline{0.316} & 0.311 & 0.329 & 0.338 \\
                     & Naïve RAG 
                     & 0.210 & 0.216 & 0.217 & 0.257 & 0.216 & 0.302 & 0.350 & 0.365 & 0.370 \\
    \textbf{ToolLinkOS} 
                     & Hybrid RAG ($\alpha=0.8$) 
                     & 0.202 & 0.208 & 0.209 & 0.265 & \underline{0.298} & 0.314 & 0.337 & 0.353 & 0.359 \\ (573 tools)
                     & Graph RAG-Tool Fusion ($k=3$), no RR
                     & \underline{0.856} & \underline{0.873} & \underline{0.873} & \underline{0.943} & \textbf{0.976} & \textbf{0.976} & \underline{0.891} & \underline{0.908} & \underline{0.908} \\ 
                     & \textbf{Graph RAG-Tool Fusion ($k=3$), w/ RR}
                     & \textbf{0.927} & \textbf{0.938} & \textbf{0.938} & \textbf{0.958} & \textbf{0.976} & \textbf{0.976} & \textbf{0.944} & \textbf{0.954} & \textbf{0.954} \\
    \hline
  \end{tabular}
  }
  \caption{Performance comparison on ToolSandbox \cite{lu2024toolsandboxstatefulconversationalinteractive} and our ToolLinkOS for different retrievers (BM25 \cite{robertson_probabilistic_2009}, Naïve RAG, Hybrid RAG (\(\alpha=0.8\)) \cite{gao2024retrievalaugmentedgenerationlargelanguage}, Graph RAG-Tool Fusion (\textit{k =} 3), and Graph RAG-Tool Fusion (\textit{k =} 3) with reranking initial vector retrieval tools. The full metrics are reported as mAP@10,20,30, recall@10,20,30, and nDCG@10,20,30.}
  \label{tab:full_retriever_comparison}
\end{table*}

\section{Instances Generation Prompt Tools}\label{sec:app_toollink_instances_prompt}
\begin{figure*}
  \centering
  \includegraphics[page=2, width=15cm,trim=0cm 0cm 0cm 0cm, clip=true]{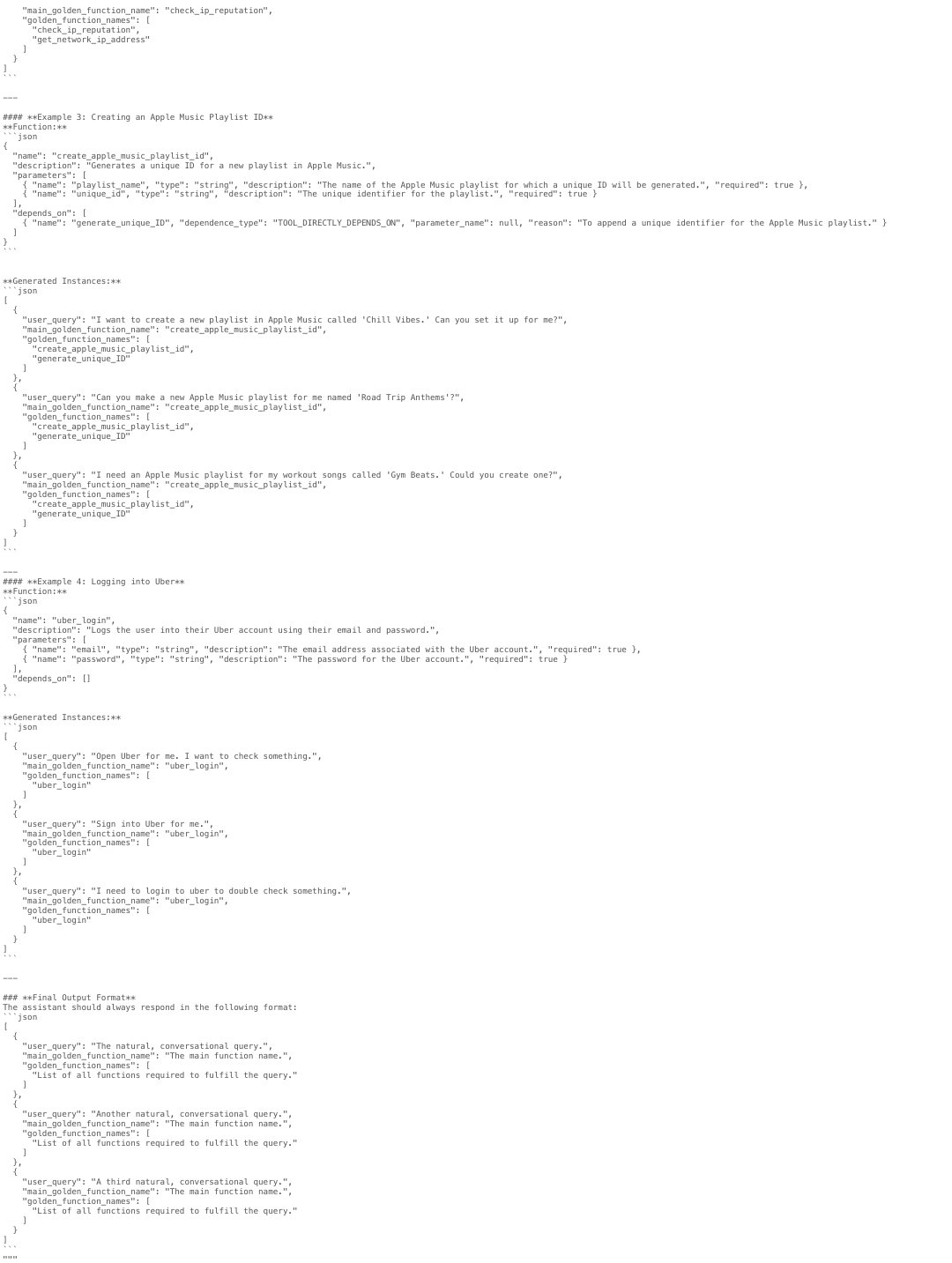} 
  \caption{LLM prompt for ToolLinkOS instances generation.}
  \label{fig:toollink_instances_prompt}
\end{figure*}

We include the LLM (gpt-4o-2024-08-06) prompt for generating ToolLinkOS instances (Figure \ref{fig:toollink_instances_prompt}).

\begin{figure*}
  \centering
  \includegraphics[page=1, width=15cm,trim=0cm 0cm 0cm 0cm, clip=true]{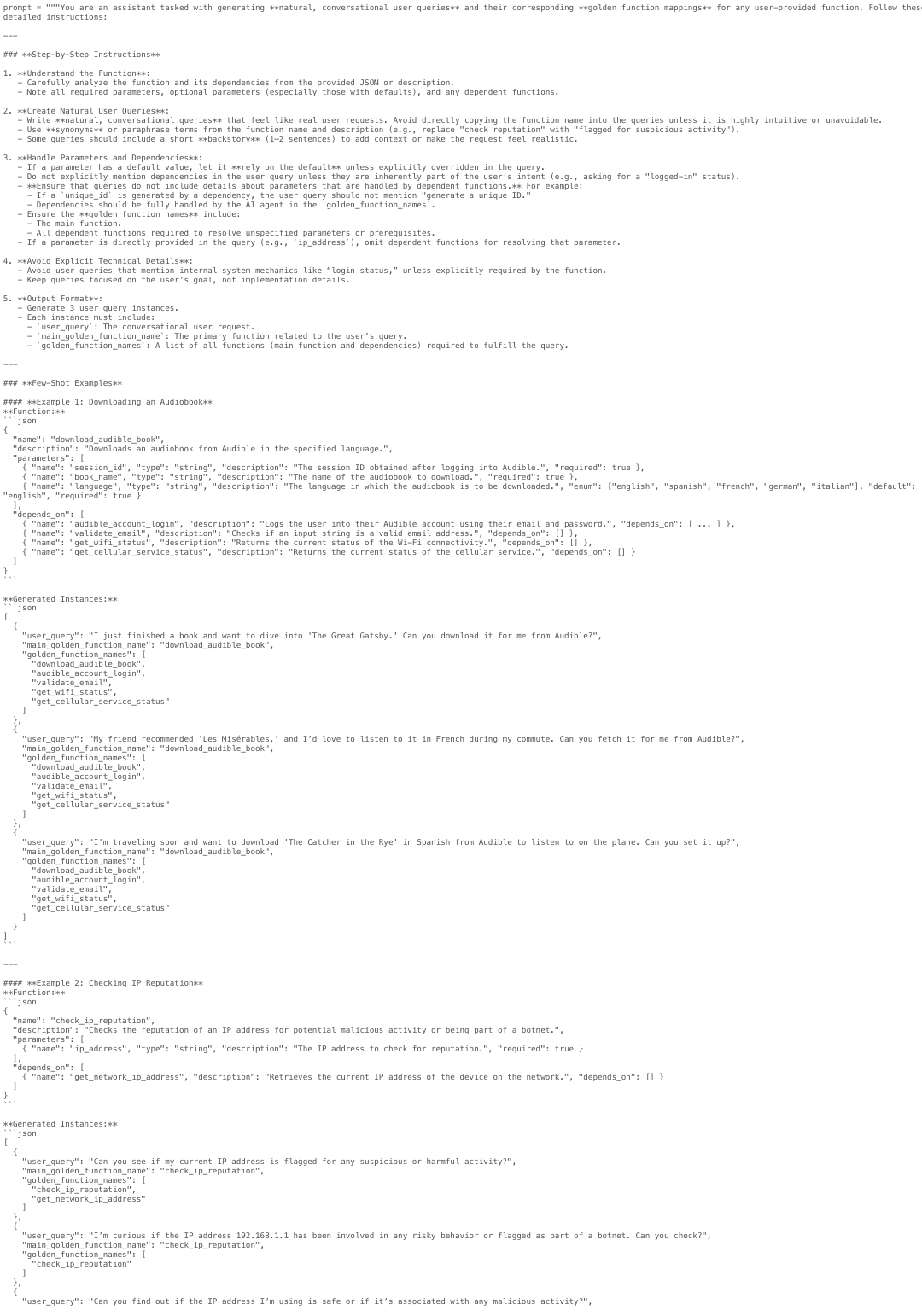}

\end{figure*}

\section{Azure AI Search Algorithm Parameters}\label{sec:app_ai_search_alg_params}
\begin{figure*}
  \centering
  \includegraphics[width=8cm]{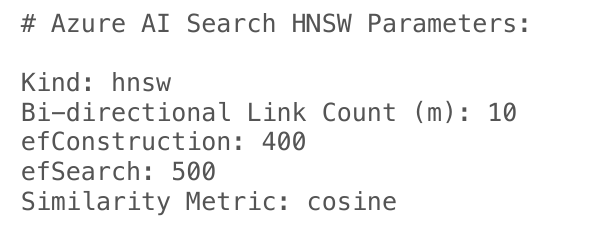} %trim=0cm 1.25cm 0cm 1.25cm, clip=true]
  \caption{Azure AI Search HNSW algorithm parameters.}
  \label{fig:azure_ai_search_alg}
\end{figure*}

The details of the Azure AI Search HNSW algorithm for similarity search are listed in Figure \ref{fig:azure_ai_search_alg}.

\section{Reranker Prompt}\label{sec:app_reranker_prompt}
\begin{figure*}[t]
  \centering
  \includegraphics[width=14cm]{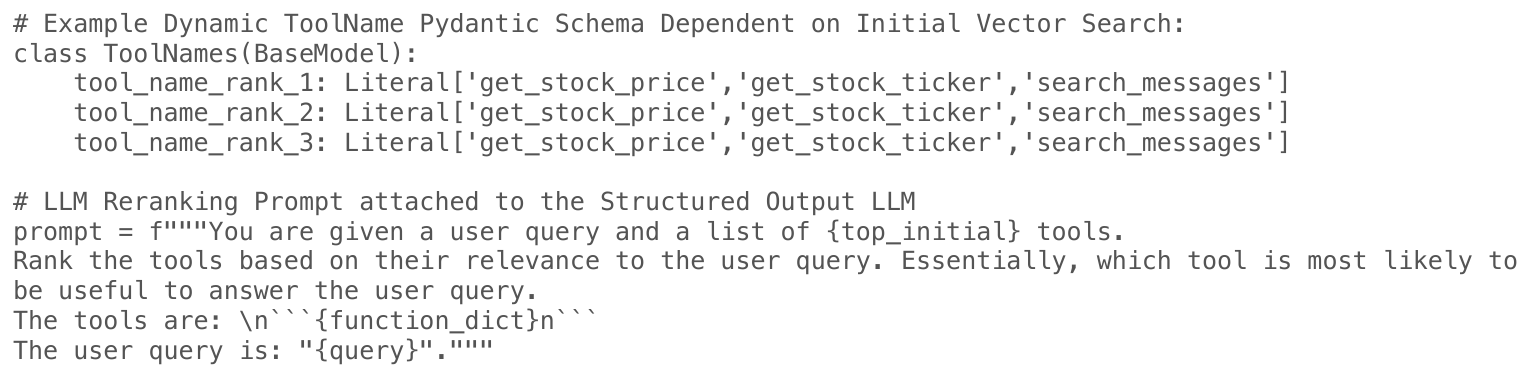} %trim=0cm 1.25cm 0cm 1.25cm, clip=true]
  \caption{Prompt for the LLM reranker of initial top-k retrieved vector tools. The structured output class is an example of the dynamic schema given to the LLM to extract the tool names in a reranked manner.}
  \label{fig:llm_reranker_prompt}
\end{figure*}

In Figure \ref{fig:llm_reranker_prompt}, the LLM reranker (gpt-4o-2024-08-06) prompt is displayed along with the structured output Pydantic schema that the LLM follows. 

\section{ToolLinkOS Industries}\label{sec:app_tool_link_os_industries}
\begin{figure*}
  \centering
  \includegraphics[width=14cm]{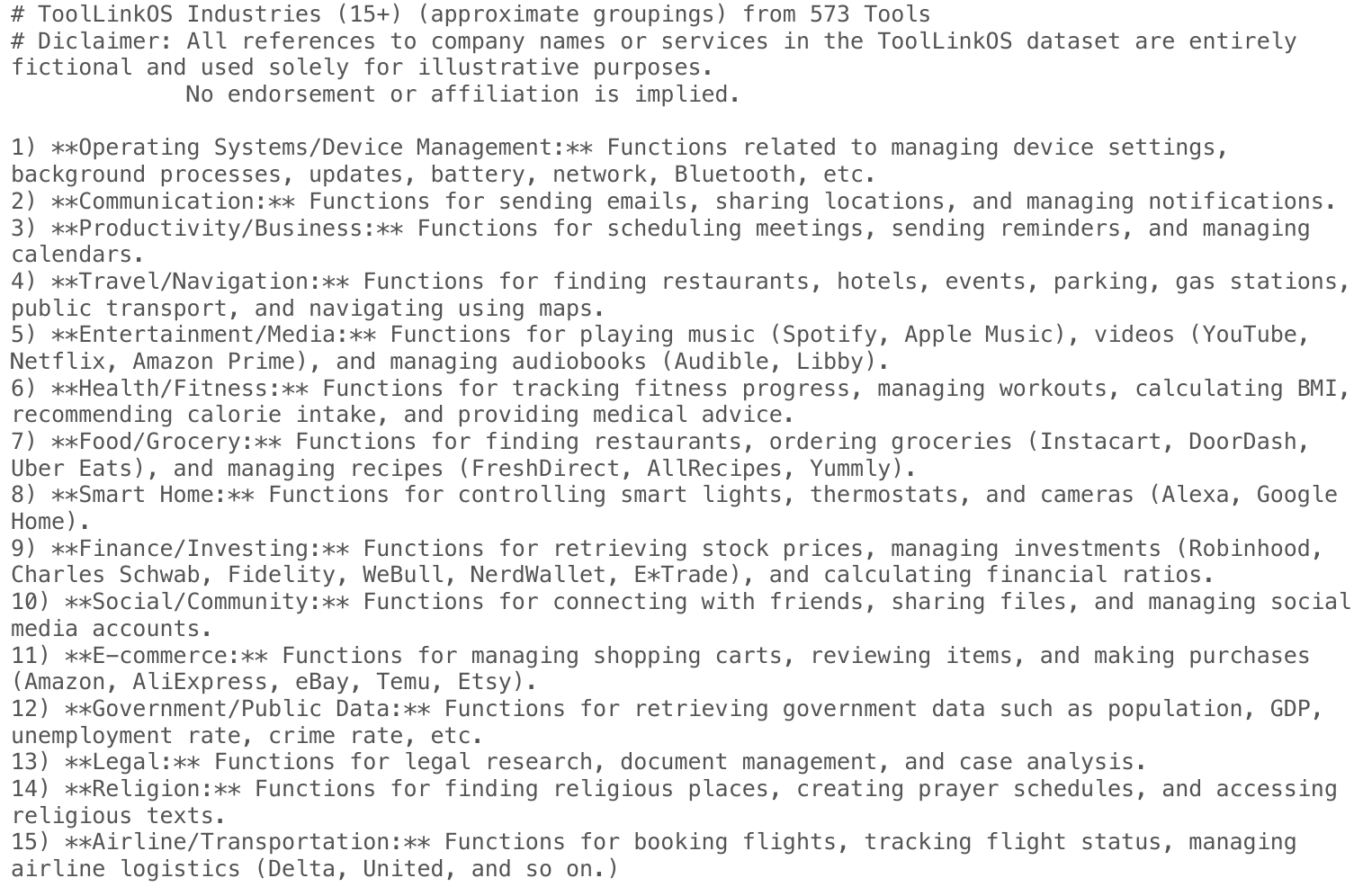} %trim=0cm 1.25cm 0cm 1.25cm, clip=true]
  \caption{Breakdown of approximately 15 industries that make up the 573 tools within the ToolLinkOS dataset. All references to company names or services in the ToolLinkOS dataset are entirely fictional and solely for illustrative purposes. No endorsement or affiliation is implied. The tools in the ToolLinkOS dataset are non-functional APIs and only consist of the knowledge graph schema JSON. }
  \label{fig:toollinkos_industries}
\end{figure*}

As described in the paper, ToolLinkOS contains 573 tools that span over 15 industries (Section~\ref{subsubsec:toollinkos_tools}). Approximate industry descriptions are displayed in Figure \ref{fig:toollinkos_industries}.

\section{ToolLinkOS Example Core Tools}\label{sec:app_toolink_core}
\begin{figure*}
  \centering
  \includegraphics[width=10cm]{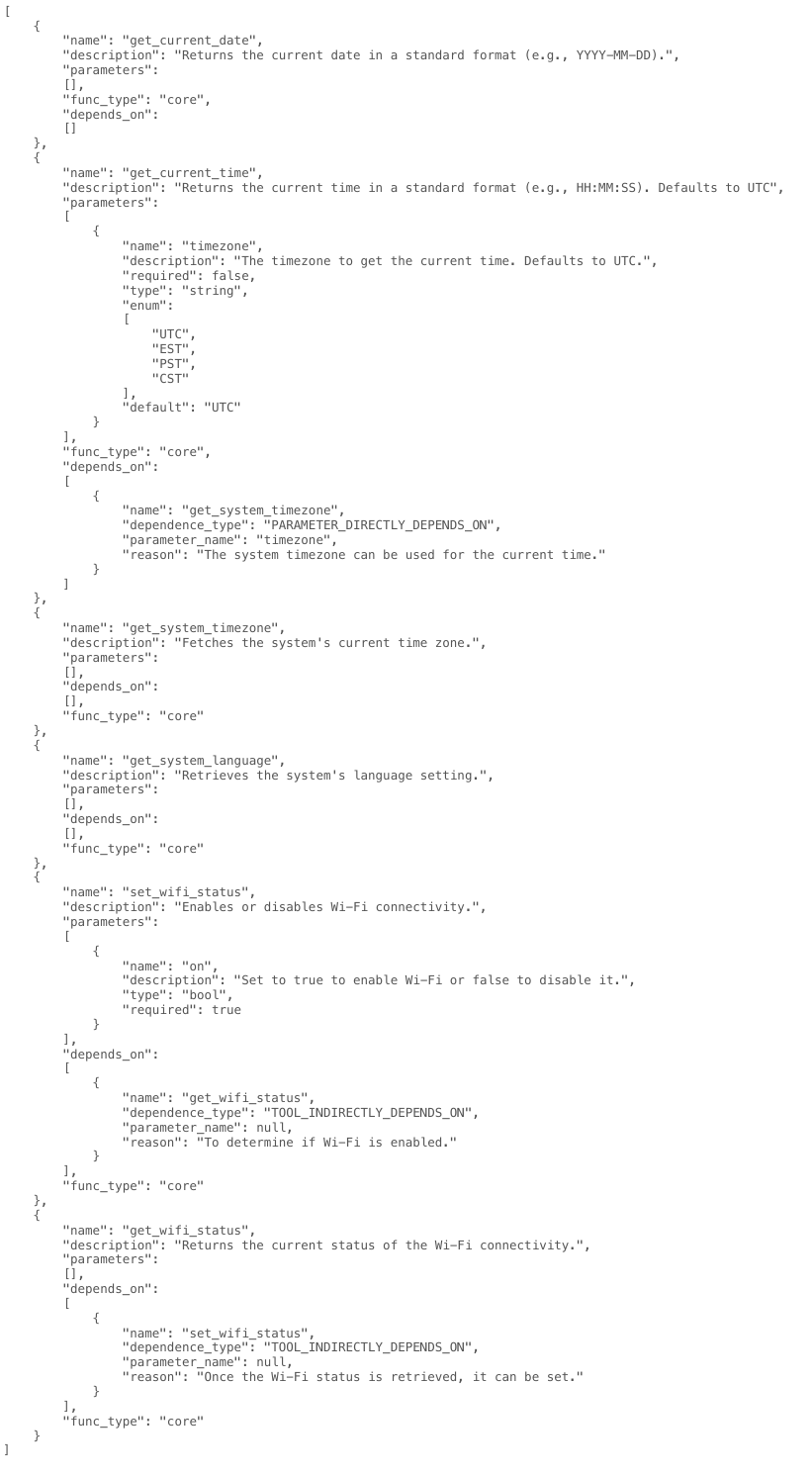} %trim=0cm 1.25cm 0cm 1.25cm, clip=true]
  \caption{Example subset of core tools from ToolLinkOS.}
  \label{fig:toollinkos_core}
\end{figure*}

In Figure \ref{fig:toollinkos_core}, a small subset of the core tools are displayed from the ToolLinkOS dataset. While the ToolLinkOS dataset has 573 tools, only 50 of them are core tools. See Section \ref{subsubsec:core_tool}.

\section{ToolLinkOS Example Regular Tools}\label{sec:app_toolink_regular}
\begin{figure*}
  \centering
  \includegraphics[page=5, width=15cm,trim=0cm 5cm 5cm 0cm, clip=true]{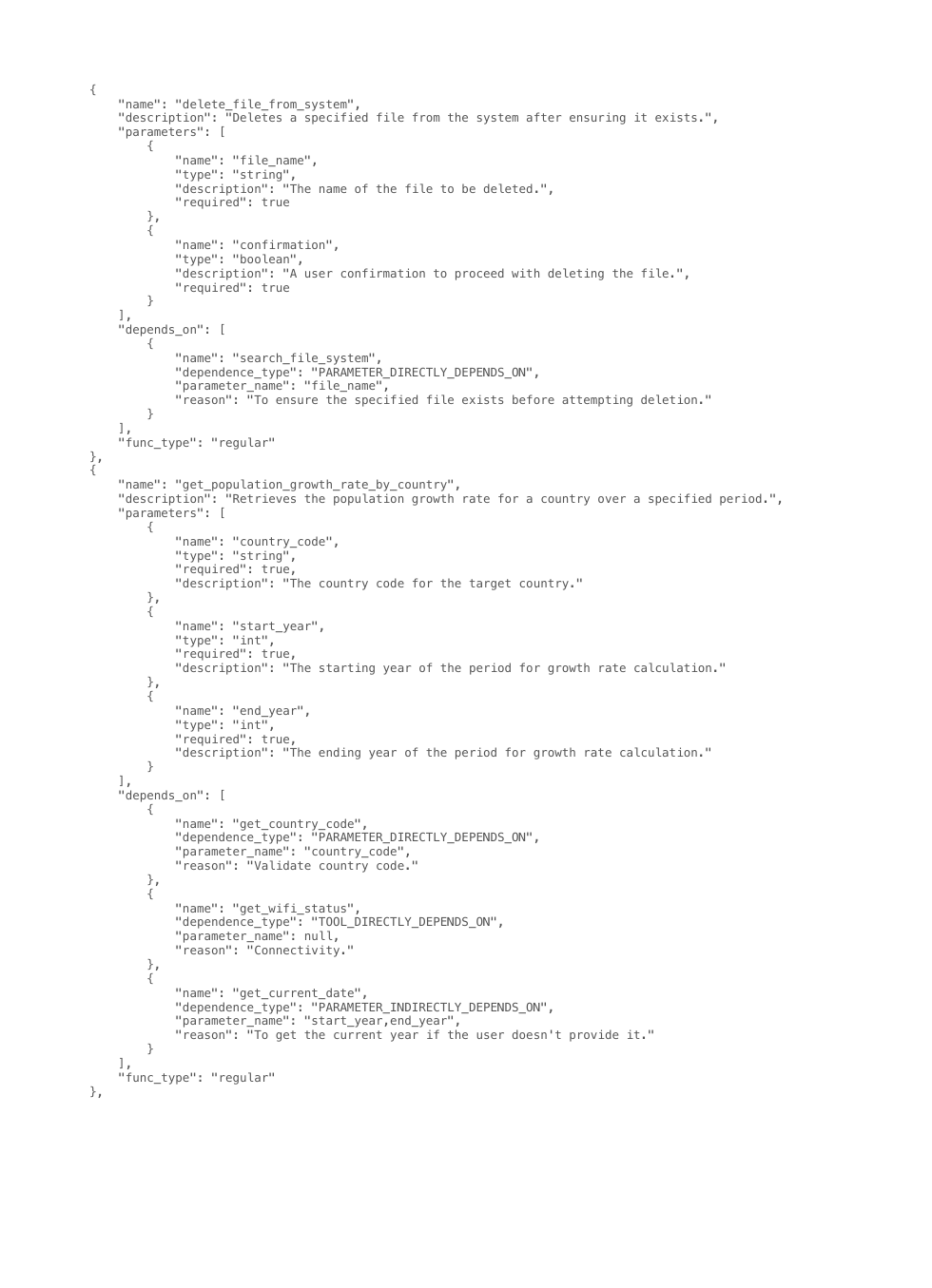}
  \caption{Example subset of regular tools from ToolLinkOS.}
  \label{fig:toollinkos_regular}
\end{figure*}

In Figure \ref{fig:toollinkos_regular}, a small subset of the regular tools are displayed from the ToolLinkOS dataset. While the ToolLinkOS dataset has 573 tools, 523 of them are regular tools. See Section \ref{subsubsec:regular_tool}.

\begin{figure*}
  \centering
  \includegraphics[page=1, width=15cm,trim=0cm 1cm 5cm 0cm, clip=true]{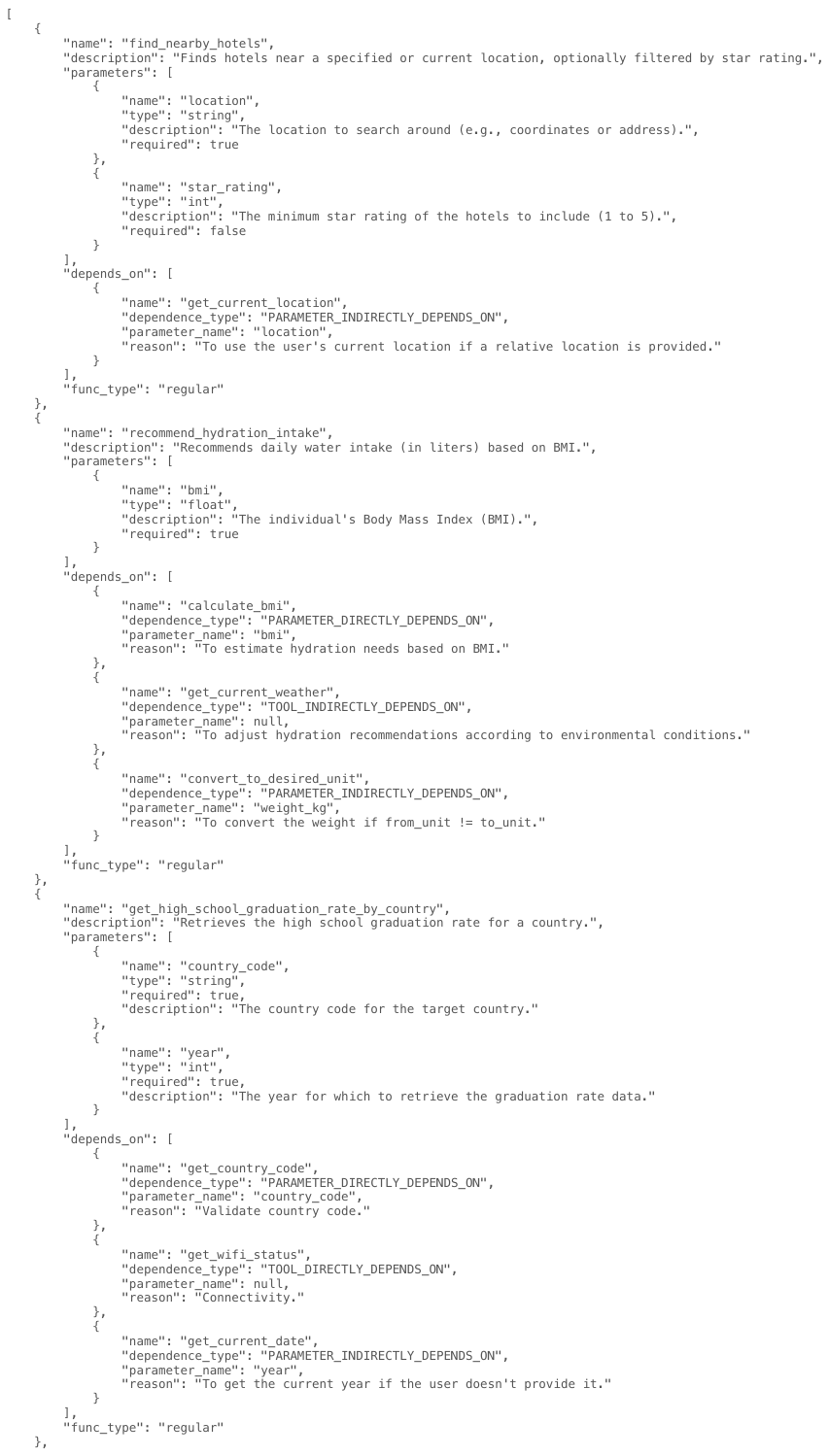} 
  %\caption{Example subset of regular tools from ToolLinkOS.}
  %\label{fig:toollinkos_core}
\end{figure*}

\begin{figure*}
  \centering
  \includegraphics[page=2, width=15cm,trim=0cm 1cm 5cm 0cm, clip=true]{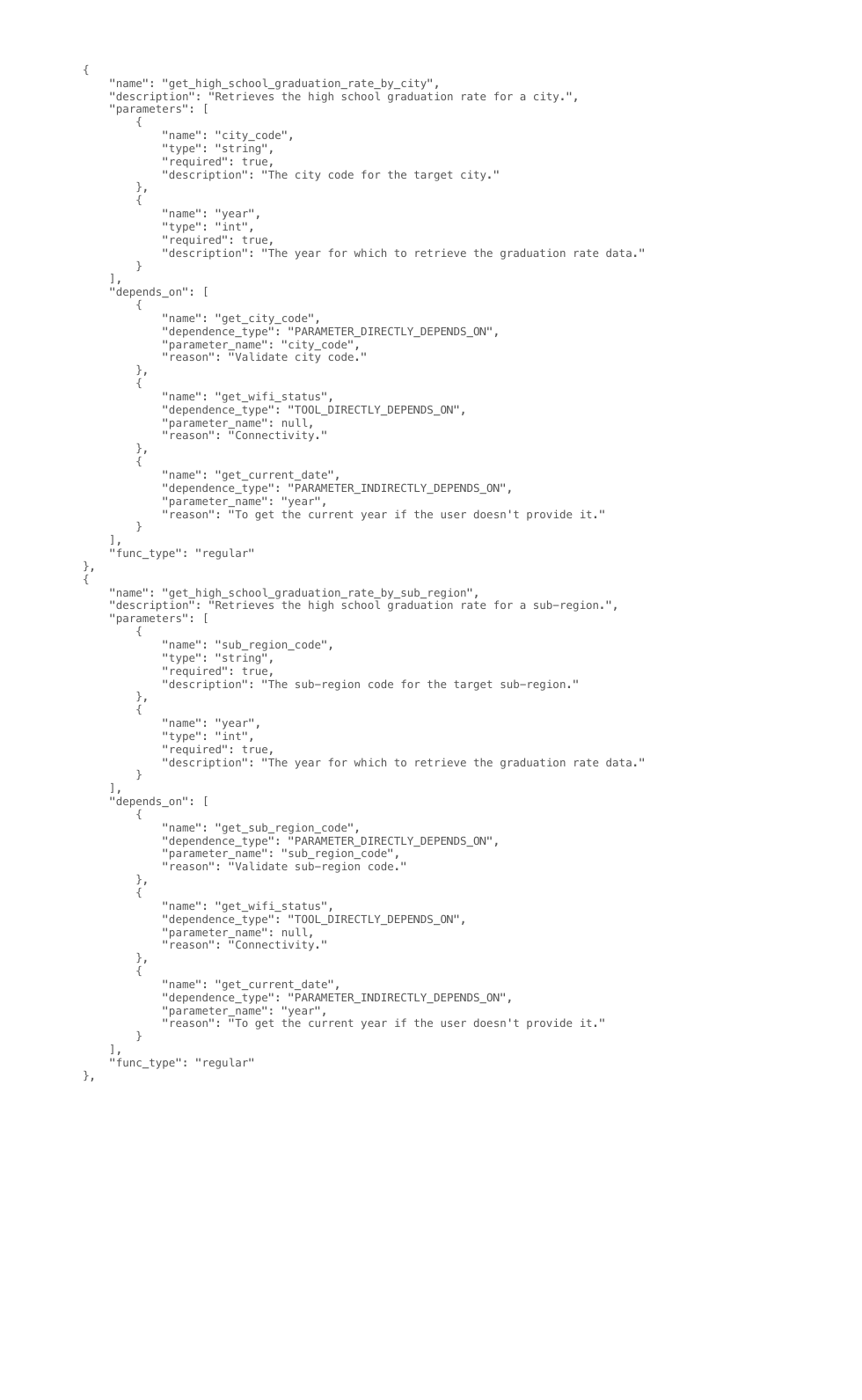} %trim=0cm 1.25cm 0cm 1.25cm, clip=true]
  %\caption{Example subset of regular tools from ToolLinkOS.}
  %\label{fig:toollinkos_core}
\end{figure*}

\begin{figure*}
  \centering
  \includegraphics[page=3, width=15cm,trim=0cm 1cm 5cm 0cm, clip=true]{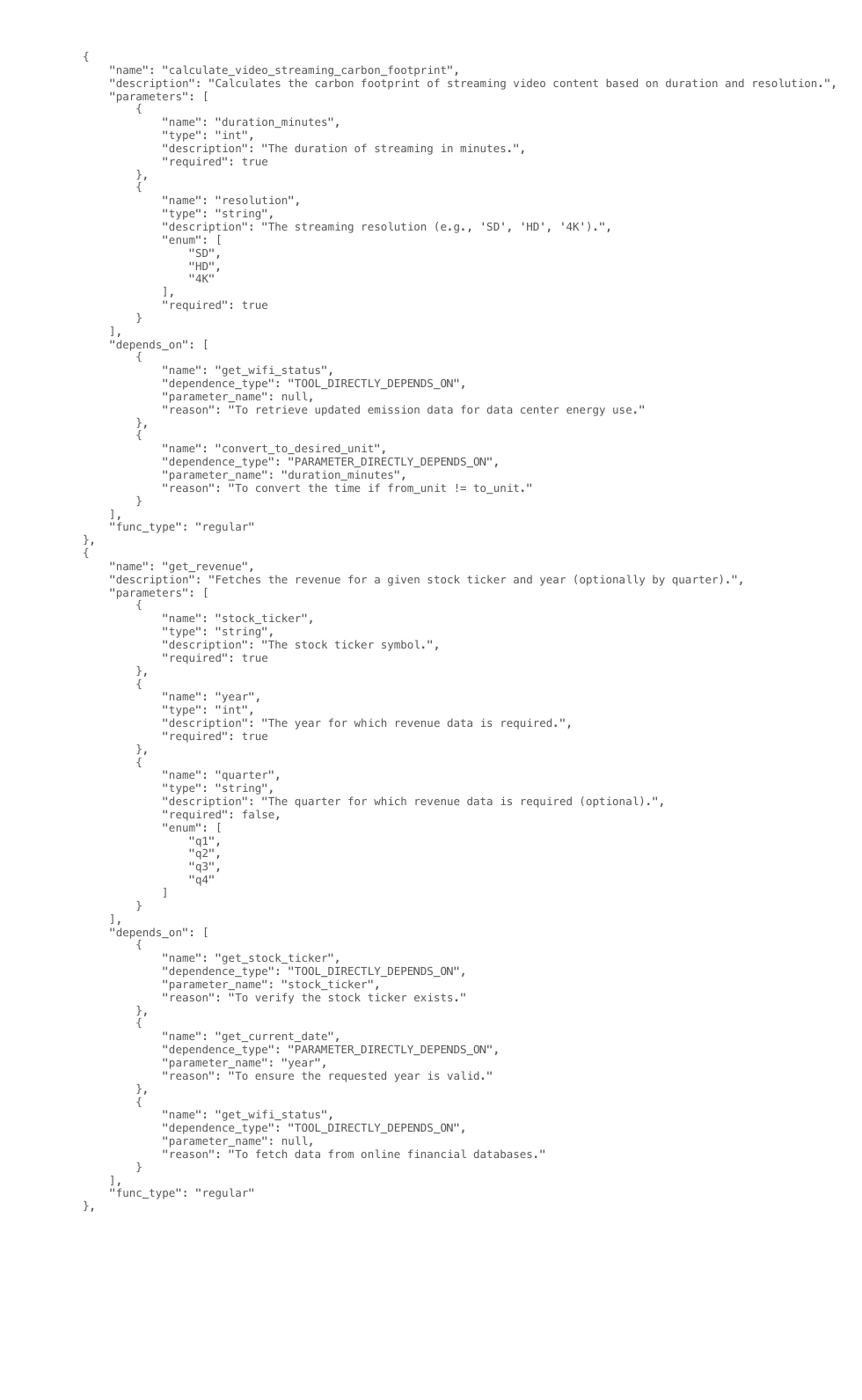} %trim=0cm 1.25cm 0cm 1.25cm, clip=true]
  %\caption{Example subset of regular tools from ToolLinkOS.}
  %\label{fig:toollinkos_core}
\end{figure*}

\begin{figure*}
  \centering
  \includegraphics[page=4, width=15cm,trim=0cm 1cm 5cm 0cm, clip=true]{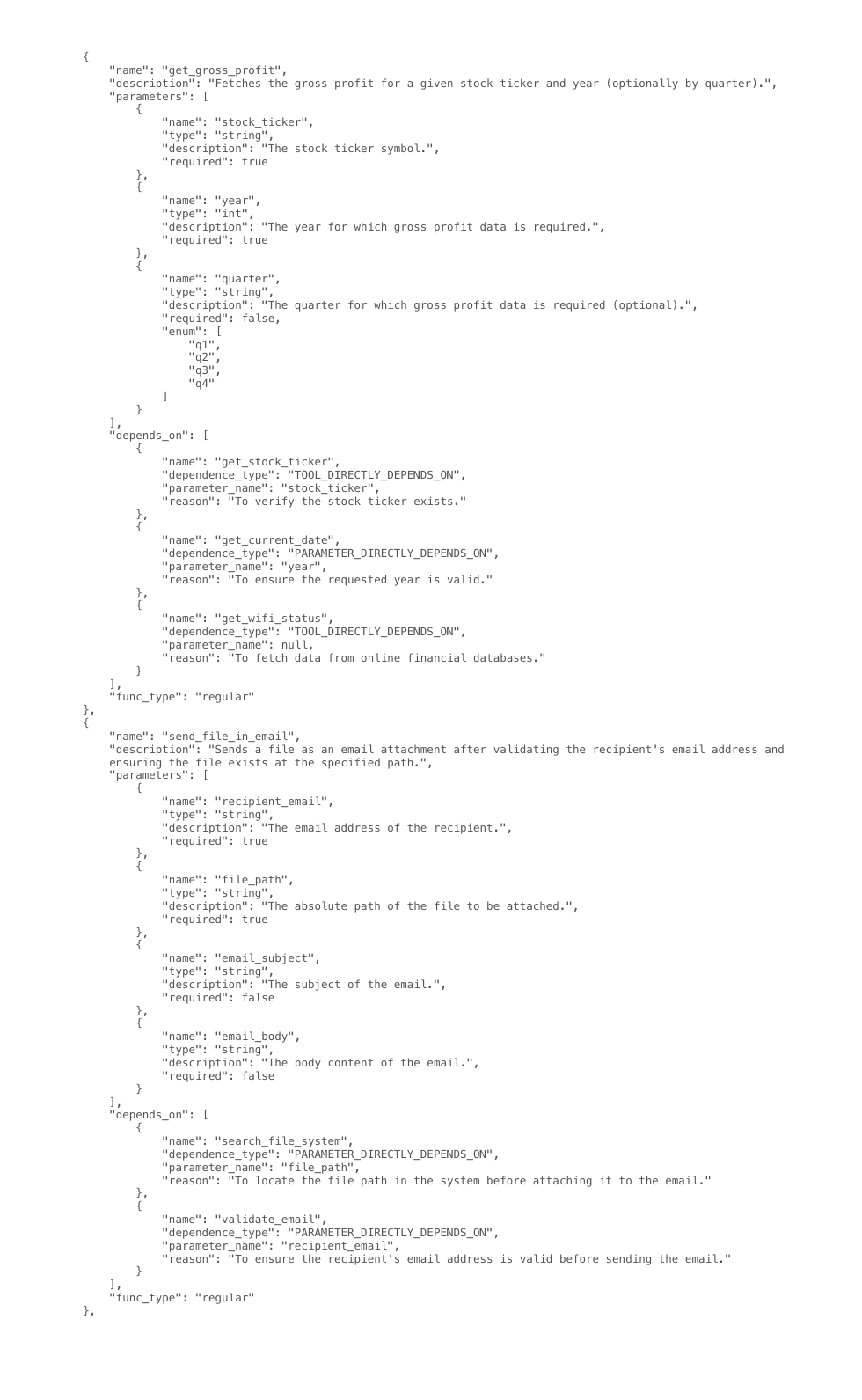} %trim=0cm 1.25cm 0cm 1.25cm, clip=true]
  %\caption{Example subset of regular tools from ToolLinkOS.}
  %\label{fig:toollinkos_core}
\end{figure*}

\begin{figure*}
  \centering
  \includegraphics[page=6, width=15cm,trim=0cm 1cm 5cm 0cm, clip=true]{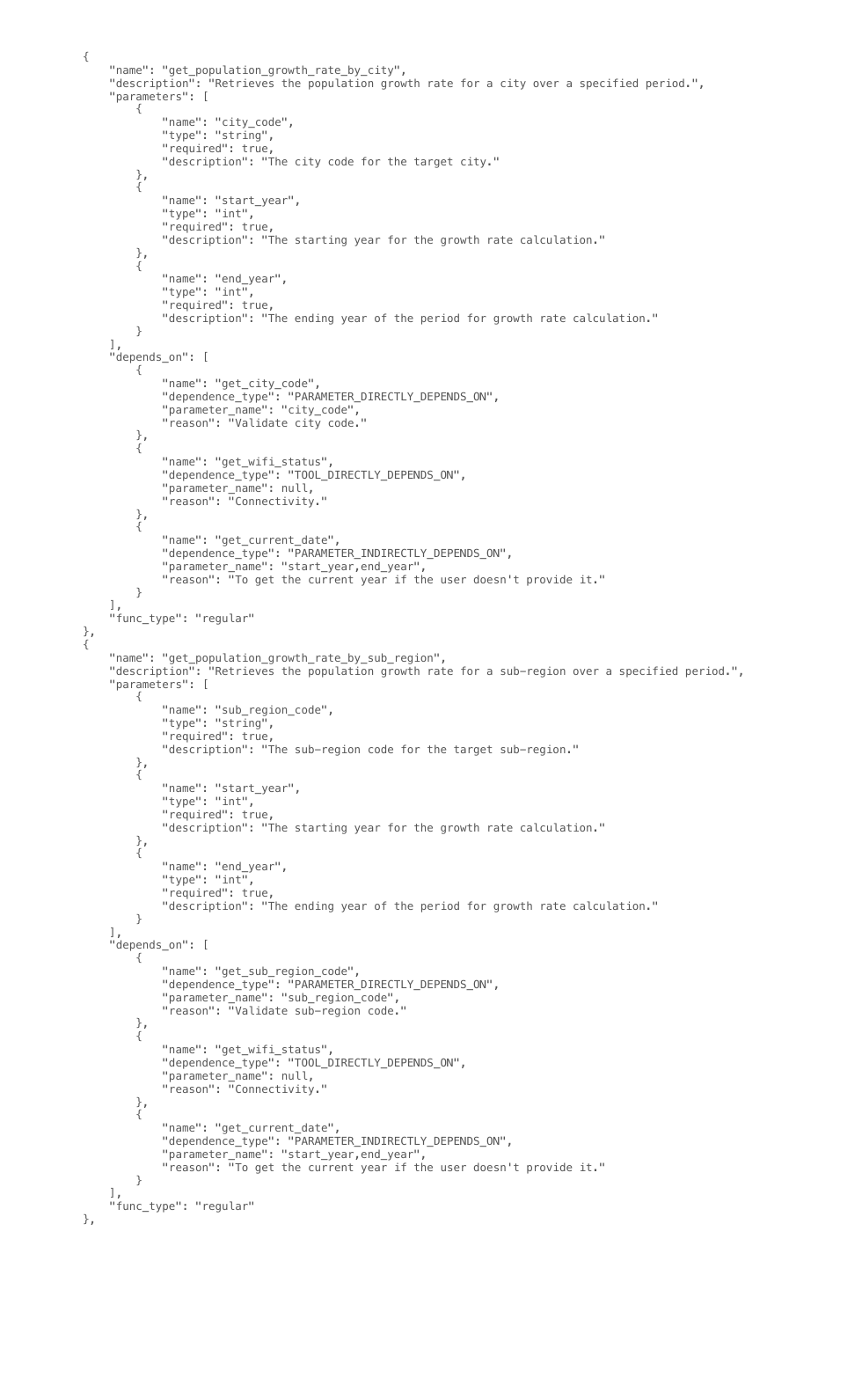} %trim=0cm 1.25cm 0cm 1.25cm, clip=true]
  %\caption{Example subset of regular tools from ToolLinkOS.}
  %\label{fig:toollinkos_core}
\end{figure*}

\begin{figure*}
  \centering
  \includegraphics[page=7, width=15cm,trim=0cm 1cm 5cm 0cm, clip=true]{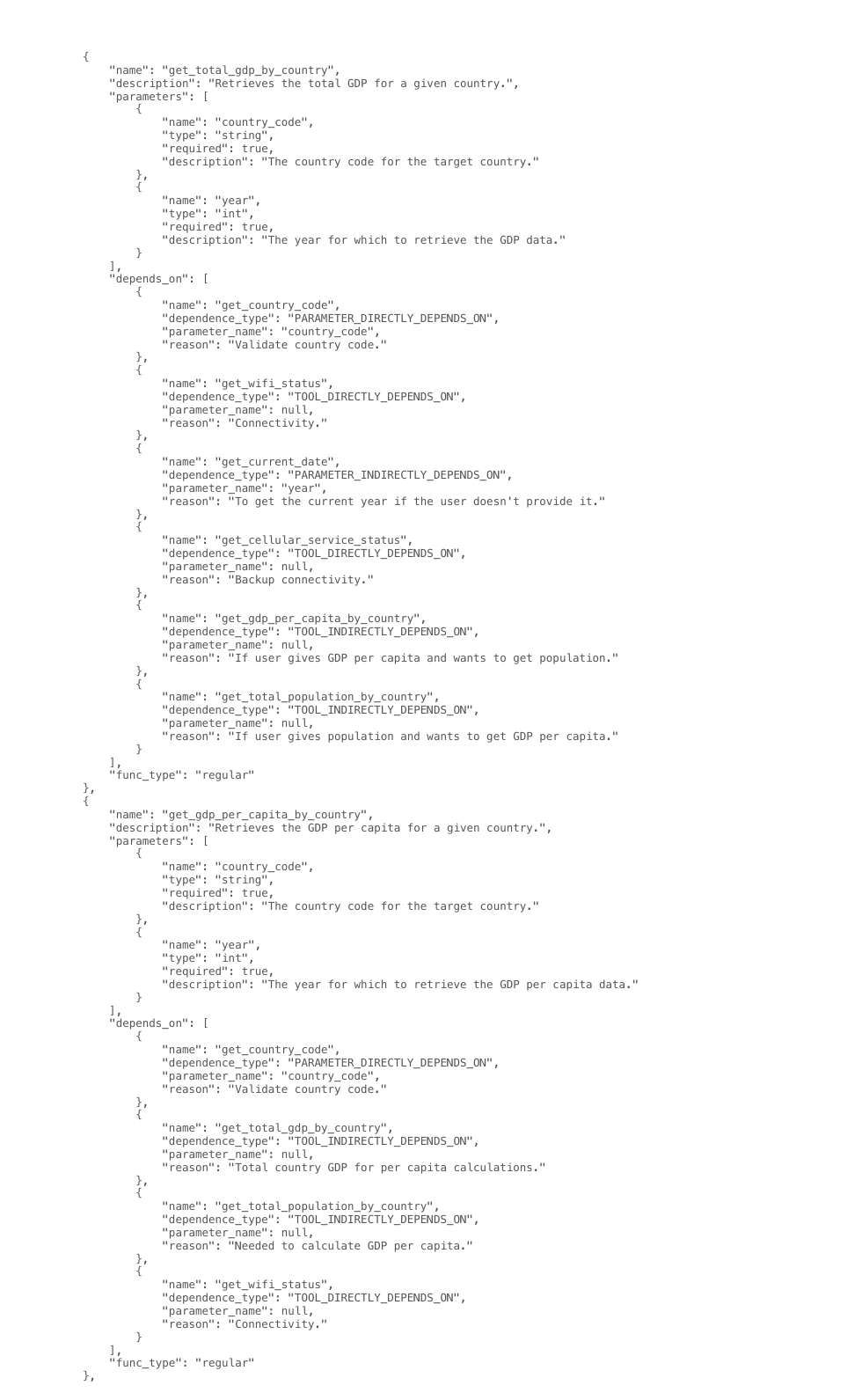} %trim=0cm 1.25cm 0cm 1.25cm, clip=true]
  %\caption{Example subset of regular tools from ToolLinkOS.}
  %\label{fig:toollinkos_core}
\end{figure*}

\begin{figure*}
  \centering
  \includegraphics[page=8, width=15cm,trim=0cm 1cm 5cm 0cm, clip=true]{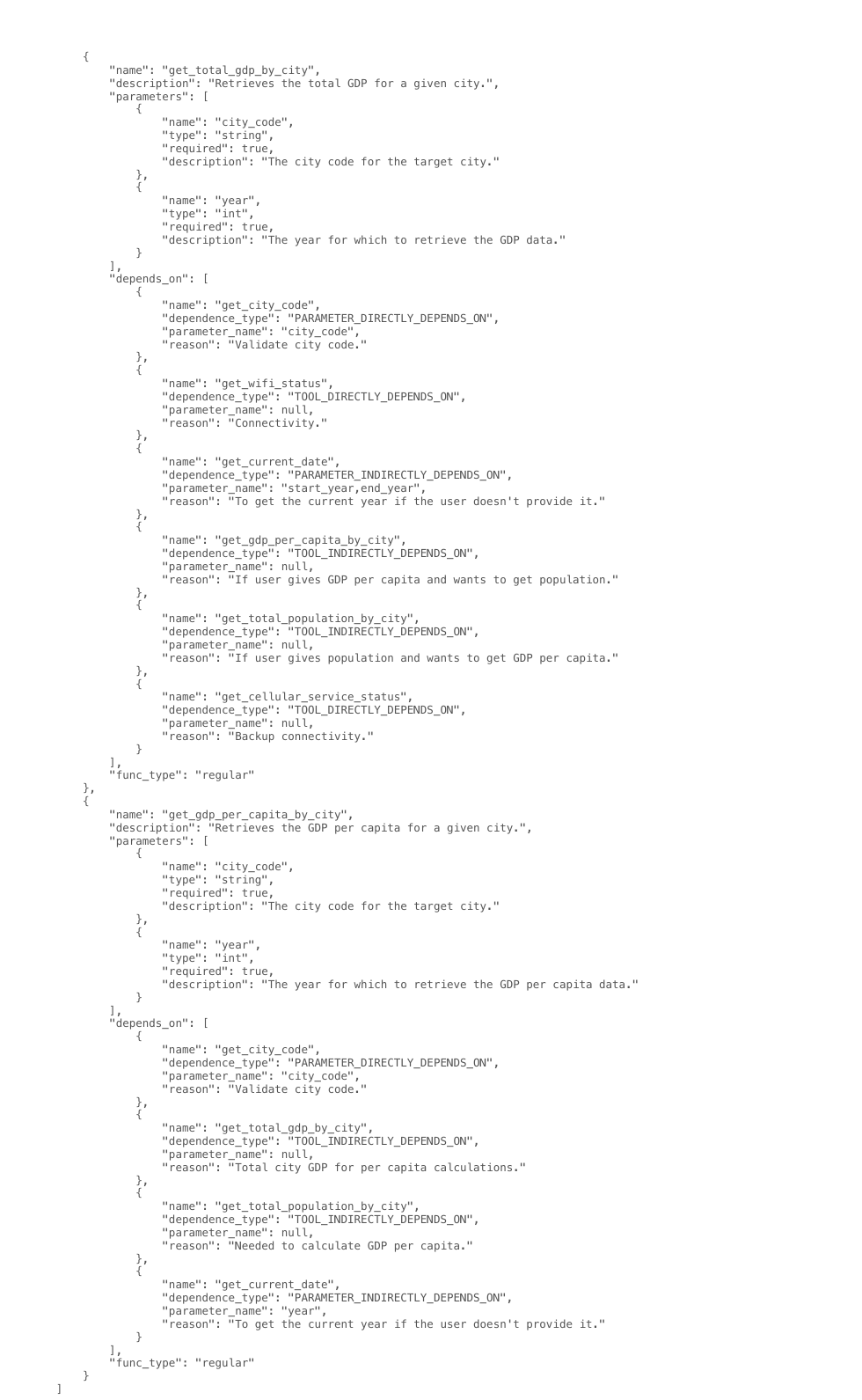} %trim=0cm 1.25cm 0cm 1.25cm, clip=true]
  %\caption{Example subset of regular tools from ToolLinkOS.}
  %\label{fig:toollinkos_regular}
\end{figure*}

\section{Tool Generation Process}\label{sec:app_tool_generation_process}
\begin{figure*}
  \centering
  \includegraphics[width=\linewidth]{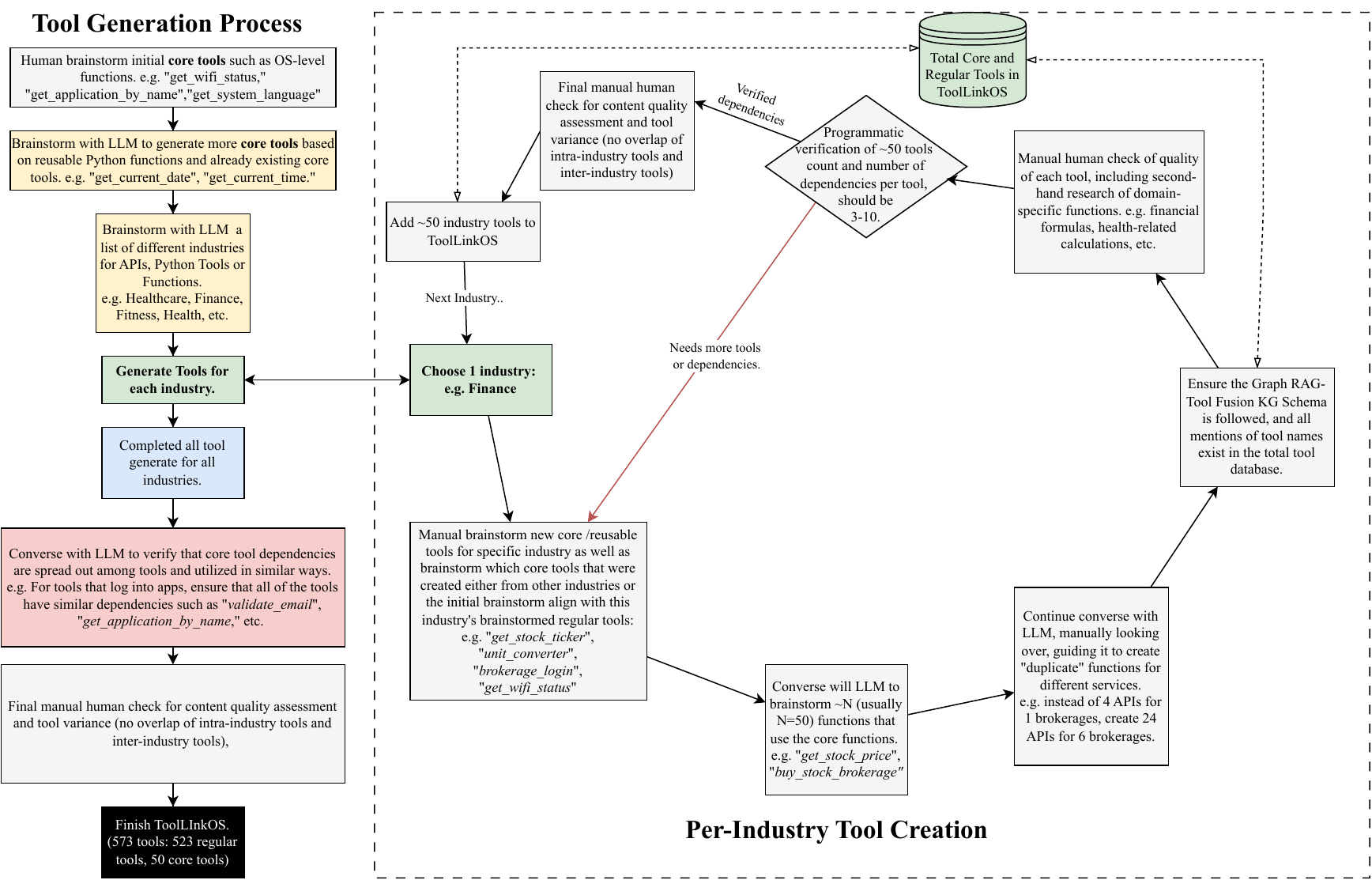} %trim=0cm 1.25cm 0cm 1.25cm, clip=true]
  \caption{ToolLinkOS tool generation process. First, brainstorm with the LLM a list of diverse industries for Python tools or APIs. Then for each industry, (1) manually brainstorm what core or reusable tools apply, (2) converse with the LLM to brainstorm $\sim$N (roughly 50) tools that use the identified core tools, (3) manually verify ways to improve tools such as "duplication of services", (4) ensure Graph RAG-Tool Fusion KG Schema is followed and no duplicate tool names exist in the dataset, (5) programmatically verify roughly 50 tools per industry and dependency count between 3-10, and (6) if confirmed, add tools to ToolLinkOS. After all tools are created, converse with LLM to ensure tool dependencies are diversified and line up with similar functioning tools. The LLMs used were gpt-4o-2024-08-06, o1-2024-12-17, o1-mini-2024-09-12.}
  \label{fig:tool_generation_process}
\end{figure*}

The Tool Generation Process involves systematically developing tools across diverse industries by leveraging human-in-the-loop conversing with LLMs. As shown in Figure~\ref{fig:tool_generation_process}, the process begins with brainstorming diverse industries for Python tools or APIs. For each industry, (1) core and reusable tools are manually identified, (2) $\sim$50 tools leveraging these cores are generated with LLM assistance, (3) tools are refined to enhance service duplication, (4) adherence to the Graph RAG-Tool Fusion KG Schema is ensured, (5) $\sim$50 tools per industry undergo programmatic verification with 3-10 dependencies, and (6) validated tools are integrated into ToolLinkOS. Finally, manually converse with LLM to ensure dependency diversity and alignment across similar tools. The LLMs used were gpt-4o-2024-08-06, o1-2024-12-17, o1-mini-2024-09-12.

\section{Supplementary Material}
The ToolLinkOS dataset is publicly available for use on GitHub (MIT License) \url{https://github.com/EliasLumer/Graph-RAG-Tool-Fusion-ToolLinkOS}.

\end{document}